\documentclass[11pt]{article}

\usepackage[preprint]{acl}

\usepackage{times}
\usepackage{latexsym}

\usepackage[T1]{fontenc}

\usepackage[utf8]{inputenc}

\usepackage{placeins}
\usepackage{float}
\usepackage{inconsolata}
\usepackage{microtype}
\usepackage{subcaption}
\usepackage{enumitem}

\usepackage[table, dvipsnames]{xcolor}
\usepackage{hyperref}
\usepackage{url}
\usepackage{graphicx}
\usepackage{algorithm}
\usepackage{amsmath}
\usepackage{colortbl}
\usepackage{multirow}
\usepackage[most]{tcolorbox}
\usepackage{booktabs}
\usepackage{algpseudocode}
\usepackage{listings}
\usepackage{bm}
\usepackage{algorithmicx}
\usepackage{pifont}
\usepackage{amssymb}
\newcommand{\gc}{\cellcolor[gray]{0.9}}
\usepackage{graphicx}

\title{
Reasoning Matters: Mitigate Hallucination in Multimodal Large Reasoning Models via Reasoning-Conditioned Preference Optimization

}

\author{
Jiawei Kong$^{1}$\thanks{Equal contribution.} \quad
Hao Fang$^{1}$\footnotemark[1] \quad
Shunxiang Liao$^{2}$ \quad
Jinyu Li$^{2}$ \\
\textbf{Bin Chen$^2$\thanks{Corresponding author.}} \quad
\textbf{Hao Wu$^{1}$} \quad
\textbf{Shu-Tao Xia$^{2}$} \quad
\textbf{Min Zhang$^{2}$} \\
\textsuperscript{1} Tsinghua Shenzhen International Graduate School, Tsinghua University \quad \\
\textsuperscript{2} Harbin Institute of Technology, Shenzhen \quad \\
\texttt{kjw25@mails.tsinghua.edu.cn, fangh25@mails.tsinghua.edu.cn}
}

\begin{document}
\maketitle

\begin{abstract}

Multimodal Large Reasoning Models introduce the reasoning paradigm, demonstrating strong capabilities on complex vision-language tasks. However, they still suffer from severe hallucinations. Existing training-based methods typically mitigate hallucinations through response-level direct preference optimization (DPO), where the Chain-of-Thought (CoT) and the final answer are treated as a monolithic output and optimized jointly. We reveal that this formulation performs similarly to answer-only optimization, suggesting that it primarily learns answer-level preference, while leaving CoT-level supervision insufficiently exploited. To address this issue, we explicitly formulate a CoT-oriented preference term and derive Reasoning-Conditioned Direct Preference Optimization (RC-DPO), which models the CoT as a condition for answer generation and contrasts the preference for the same preferred answer under different CoT conditions, promoting answer-supportive reasoning chain alignment. To further improve optimization, we introduce a reasoning-enhanced preference data generation strategy that employs Monte Carlo Tree Search to discover visually grounded and logically consistent CoTs as positive samples, and attention-guided CoT token pruning to construct negative ones. Extensive experiments across various models and benchmarks show that RC-DPO effectively mitigates hallucinations and improves the reliability of the multimodal reasoning process.

\end{abstract}

\section{Introduction}

Large Reasoning Models (LRMs), such as OpenAI o3 and DeepSeek-R1 \cite{guo2025deepseek}, have recently achieved remarkable progress on various tasks by explicitly generating intermediate reasoning steps before producing final responses \cite{cobbe2021training, chen2021evaluating}. Building upon this, Multimodal Large Reasoning Models (MLRMs) \cite{meng2025mm, yang2025r1, leng2025mmr1, wang2026sota} extend such reasoning capabilities to vision-language scenarios. Further enhanced by post-training techniques such as reinforcement learning, these models have achieved substantial gains on challenging multimodal tasks, including visual mathematics \cite{lu2024mathvista}, scientific question answering \cite{lu2022learn}, and chart understanding \cite{masry2022chartqa}.

\begin{figure}[!t]
\begin{center}
\includegraphics[width=\linewidth]{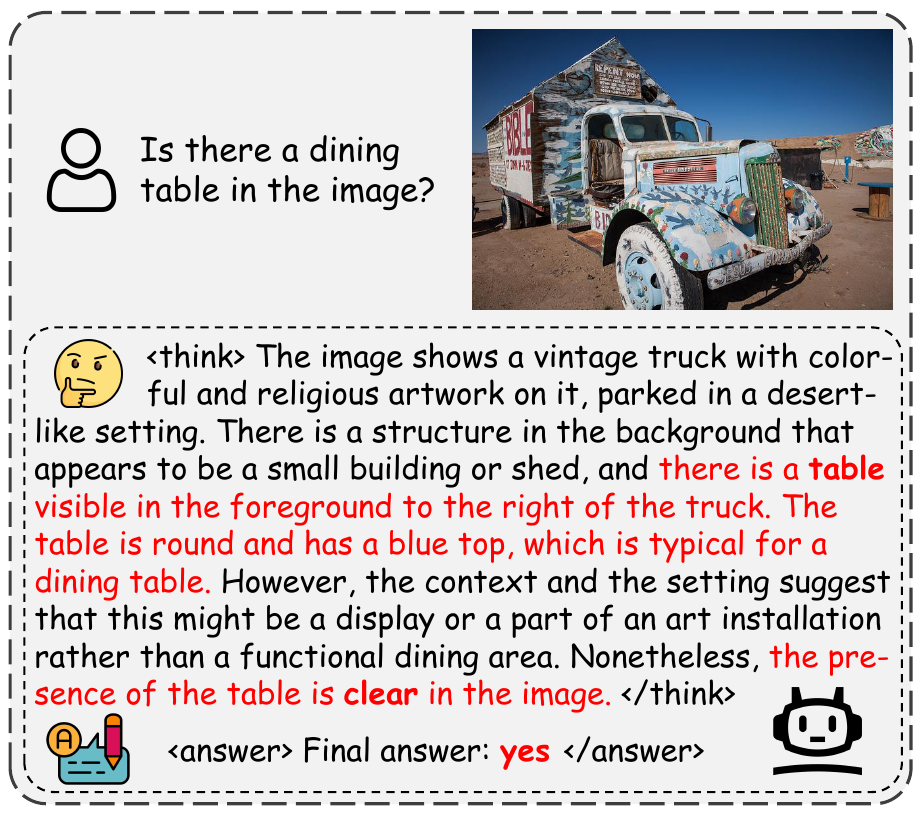}
\end{center}
\caption{A visual question answering example generated by MM-Eureka-7B \cite{meng2025mm}, where the hallucinated content is highlighted in \textcolor{red}{red}.}
\label{fig:intro}
\end{figure}

Despite their strong performance, recent studies \cite{xu2026more, fang2026seeing} have shown that MLRMs remain prone to hallucinations, generating content that appears plausible yet is inconsistent with the visual input. Unlike conventional Multimodal Large Language Models (MLLMs), hallucinations in MLRMs can arise not only in the final answer but also within the intermediate Chain-of-Thought (CoT), as illustrated in Figure~\ref{fig:intro}. 
This poses a unique challenge for hallucination mitigation: unsupported visual evidence, such as nonexistent objects or relations, may first emerge in the CoT and then propagate to the final answer through subsequent reasoning steps. Therefore, mitigating hallucinations in MLRMs requires moving beyond answer-level correction toward the explicit alignment of intermediate reasoning chains.

Recent studies have explored post-alignment methods for mitigating hallucinations in MLRMs, with Direct Preference Optimization (DPO) commonly employed to align model outputs with curated preference pairs \cite{yu2025rlaif, yang2025mitigating}. However, these methods typically construct preference pairs over complete responses, where the CoT and final answer are concatenated and jointly optimized under a single response-level preference signal. 
While such coarse-grained supervision can reduce hallucinations to some extent, it does not explicitly distinguish CoT-level alignment from answer-level alignment. As a result, the optimization may be \textit{biased toward answer-level shortcuts}, leaving the reasoning chain insufficiently optimized and limiting the overall effectiveness of hallucination mitigation. This raises a natural question: \textit{How can we move beyond response-level preference optimization and explicitly align reasoning chains with the correct answer?}

To address this, we propose \textbf{R}easoning-\textbf{C}onditioned \textbf{D}irect \textbf{P}reference \textbf{O}ptimization (RC-DPO), a fine-grained preference optimization framework that explicitly models the relationship between reasoning chains and final answers. Instead of only comparing complete responses, RC-DPO treats the CoT as a condition for answer generation and contrasts the likelihood of the same preferred answer under different CoT conditions. Specifically, given a multimodal input and a preferred answer, RC-DPO encourages the model to assign higher preference to the answer when conditioned on a visually grounded and logically valid CoT than on a hallucinated or corrupted CoT. 
By isolating the effect of the reasoning chain, this controlled comparison enables the model to better distinguish quality differences between reasoning chains and provides a direct signal for learning answer-supportive CoT preferences.

To support this objective, we further construct reasoning-conditioned preference data by searching for high-quality positive reasoning chains with Monte Carlo Tree Search (MCTS), and generating negative reasoning chains through an attention-guided CoT token pruning strategy that removes visually salient CoT tokens. 
By combining response-level preference optimization with reasoning-conditioned preference learning, RC-DPO explicitly aligns intermediate reasoning chains with the answers they support, enabling more faithful multimodal reasoning.

In summary, our contributions are as follows:
\begin{itemize}

\item We revisit existing DPO-based hallucination mitigation methods for MLRMs and reveal that their coarse-grained supervision may leave CoT insufficiently optimized.

\item We propose RC-DPO, a fine-grained preference optimization framework that aligns reasoning chains by contrasting the same preferred answer under different CoT conditions.

\item We further introduce a reasoning-conditioned preference data generation pipeline with MCTS-based positive sampling and attention-guided negative construction.

\item  Experimental results on four MLRMs across nine benchmarks demonstrate the effectiveness of RC-DPO in mitigating hallucinations and improving multimodal reasoning ability.

\end{itemize}

\section{Observation and Motivation}
\label{sec:motivation}

In this section, we present the motivation behind our RC-DPO for mitigating hallucinations in MLRMs. We first provide evidence of insufficient CoT optimization from two complementary perspectives, \textit{i.e.}, training dynamics and inference-time hallucination decomposition. 
We further highlight the necessity of CoT-level hallucination mitigation by analyzing the conditional dependency between CoT and answer hallucinations.

\begin{figure*}[htbp]
\begin{center}
\includegraphics[width=\linewidth]{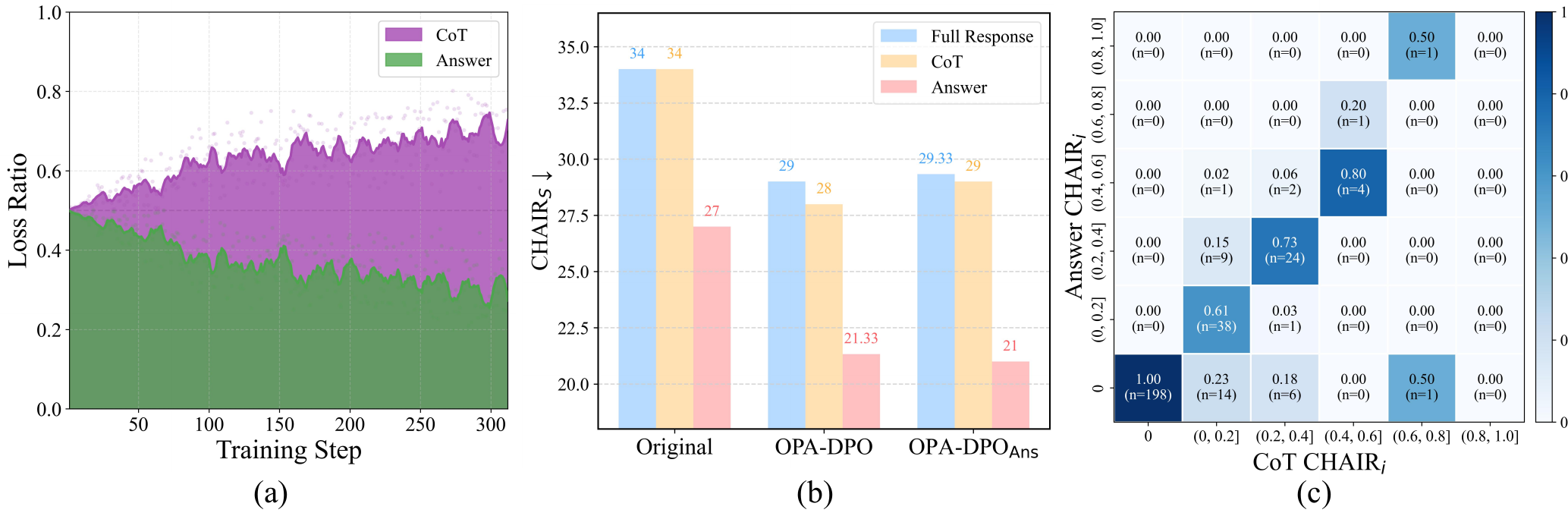}
\end{center}
\vspace{-0.5em}
\caption{Motivating observations for RC-DPO. (a) Loss-ratio curves of CoT and answer segments during OPA-DPO training. (b) Segment-level CHAIR$_S$ scores of the original model, OPA-DPO, and the answer-only variant OPA-DPO$_{\text{Ans}}$. (c) Conditional distribution between CoT CHAIR$_i$ scores and answer CHAIR$_i$ scores.}
\label{fig:motivation}
\end{figure*}

\subsection{Answer Bias in Response-level DPO}

\textbf{Observation.}
We analyze the training dynamics of OPA-DPO~\cite{yang2025mitigating} by decomposing each response into CoT and answer spans and tracking their loss ratios during training. As shown in Figure~\ref{fig:motivation} (a), the policy-reference preference gap grows faster on the answer segment than on CoT, indicating that answer-level signals are optimized more rapidly during response-level DPO.

To further examine this bias, we implement an answer-only variant, $(c_w,a_w) \succ (c_w,a_l)$, where the same CoT is shared and only the final answer differs. This removes CoT-level preference from the comparison and provides only answer-level supervision. As shown in Figure~\ref{fig:motivation} (b), its CHAIR$_S$ score is close to that of OPA-DPO.

\textbf{Insights.}
These results suggest that response-level DPO may be biased toward answer-level optimization, with much of its gain arising from answer-level preference learning, while CoT preference remains insufficiently captured. 
This motivates a more fine-grained objective that explicitly separates CoT alignment from answer-level alignment to reduce reliance on shortcut learning.

\begin{figure*}[htbp]
\begin{center}
\includegraphics[width=\linewidth]{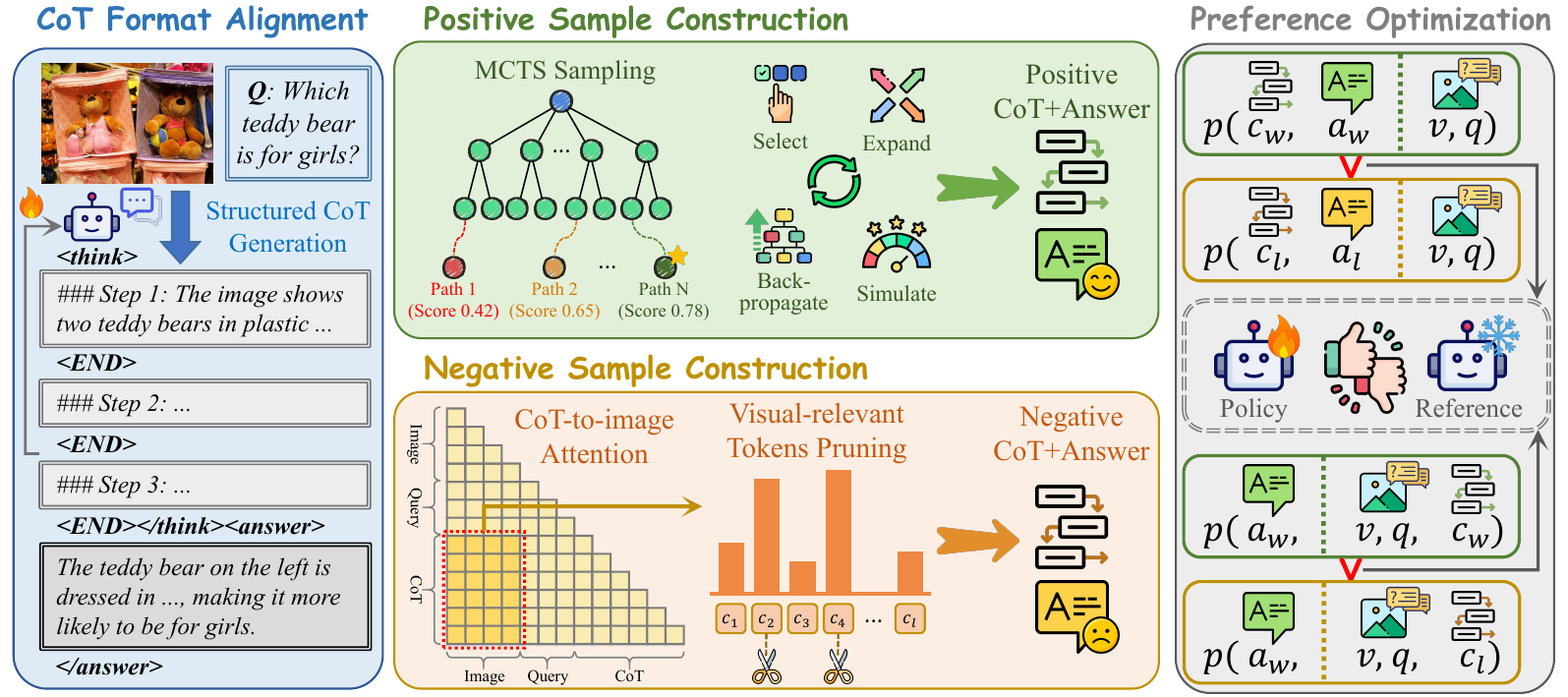}
\end{center}
\vspace{-0.5em}
\caption{
Overview of RC-DPO. We first construct SFT data to align the CoT format. Then, we construct positive and negative reasoning samples through MCTS sampling and attention-guided token pruning, respectively. Finally, we optimize the policy model with both response-level and reasoning-conditioned preference objectives.
}
\label{fig:pipeline}
\vspace{-0.5em}
\end{figure*}

\subsection{CoT as the Hallucination Bottleneck}

\textbf{Observation.}
We evaluate CHAIR$_S$ on OPA-DPO outputs, including the complete response and its decomposed CoT and answer segments. As shown in Figure~\ref{fig:motivation} (b), although OPA-DPO reduces hallucinations compared with the original model, CoT hallucination remains consistently higher than answer hallucination, with the full-response score staying closer to the CoT score.

\textbf{Insights.}
This indicates that residual full-response hallucinations are largely dominated by hallucinated reasoning chains, making CoT the key bottleneck. Therefore, improving the final answer alone is insufficient, and effective hallucination mitigation requires explicit CoT-level alignment.

\subsection{CoT-Conditioned Answer Hallucination}
\textbf{Observation.}
We further analyze the conditional probability of answer hallucination given CoT hallucination based on segment-level CHAIR$_i$ scores.
As shown in Figure~\ref{fig:motivation} (c), answers remain hallucination-free when no hallucinated content appears in the CoT, as indicated by the first column. In contrast, answer hallucinations become more frequent once hallucinations appear in the CoT, revealing a conditional dependency between CoT hallucination and answer hallucination.

\textbf{Insights.}
This conditional dependence suggests that answer hallucination is closely associated with hallucinations in the reasoning chain. Unsupported visual claims introduced during reasoning may propagate to the final answer, making answer-level correction alone insufficient. Therefore, effective hallucination mitigation should explicitly align reasoning chains, ensuring that CoTs provide reliable support for the preferred answer.

\section{Methodology}

\subsection{Preliminaries}

Before delving into the proposed reasoning-conditioned algorithm, \textit{i.e.}, RC-DPO, we first revisit the generation paradigm of MLRMs and the formulation of standard DPO, which provide the foundation for our subsequent derivations.

\paragraph{Multimodal Reasoning with CoT.}
Given a multimodal input $x=(v,q)$, where $v$ denotes the image and $q$ denotes the user prompt, an MLRM $\pi_\theta$ generates a response $y$ consisting of an intermediate reasoning chain $c$ and a final answer $a$. 
Under the standard autoregressive formulation, the likelihood of the response can be factorized as
\begin{equation}
\pi_\theta(y|x)=\pi_\theta(c,a|x) =\pi_\theta(c|x)\pi_\theta(a|x,c).
\end{equation}
This decomposition shows that response generation involves two related but distinct components, \textit{i.e.}, generating the reasoning chain and producing the final answer conditioned on that chain.

\paragraph{Response-level DPO.}
Standard DPO optimizes a policy model with preference pairs. Given an input $x$, a preferred response $y_w$, and a rejected response $y_l$, the DPO objective is formulated as
\begin{equation}
\small
\mathcal{L}_{\mathrm{DPO}}
=
-\log \sigma \left(
\beta
\left[
\log \frac{\pi_\theta(y_w|x)}{\pi_{\mathrm{ref}}(y_w|x)}
-
\log \frac{\pi_\theta(y_l|x)}{\pi_{\mathrm{ref}}(y_l|x)}
\right]
\right),
\end{equation}
where $\pi_{\mathrm{ref}}$ indicates a fixed reference model and $\beta$ controls the strength of regularization.

For MLRMs, existing methods typically construct preference pairs over complete responses, \textit{i.e.}, $y_w=(c_w,a_w)$ and $y_l=(c_l,a_l)$. 
Therefore, the response-level DPO can be formulated as
\begin{equation}
\begin{aligned}
\mathcal{L}_{\mathrm{DPO}}
= -\log \sigma \Bigg(
&\beta \Bigg[
\log \frac{\pi_\theta(c_w, a_w|x)}
{\pi_{\mathrm{ref}}(c_w, a_w|x)} \\
&-
\log \frac{\pi_\theta(c_l, a_l|x)}
{\pi_{\mathrm{ref}}(c_l, a_l|x)}
\Bigg]
\Bigg).
\end{aligned}
\end{equation}
As revealed in Section \ref{sec:motivation}, such response-level preference signals remain coarse-grained, as they do not explicitly distinguish CoT-level alignment from answer-level alignment, leading to answer shortcuts and insufficient CoT optimization.


\subsection{The Proposed RC-DPO}
Motivated by the above analysis, we propose RC-DPO to explicitly align reasoning chains by modeling the CoT as a condition for answer generation. 
The overall pipeline is illustrated in Figure~\ref{fig:pipeline}.

\paragraph{CoT Format Alignment.}
We first perform supervised fine-tuning (SFT) to align the model with a structured CoT format. This stage not only serves as a warm-up phase for subsequent DPO training, but also adapts the model to generate well-structured reasoning trajectories suitable for MCTS-based positive sample construction.
Specifically, given an input $x=(v,q) \in \mathcal{D}$, we prompt the model with few-shot demonstrations, as illustrated in Appendix~\ref{appendix:prompt}, to generate a step-wise CoT $c$ followed by a final answer $a$. The resulting triples form the SFT dataset $\mathcal{D}_{SFT}$, on which the model is trained with the standard supervised objective:
\begin{equation}
\mathcal{L}_{\mathrm{SFT}}
=
-\mathbb{E}_{(x,c,a)\sim \mathcal{D}_{\mathrm{SFT}}}
\left[
\log \pi_\theta(c,a|x)
\right].
\label{eq:sft}
\end{equation}
After this warm-up stage, the model $\pi_{\mathrm{SFT}}$ can produce structured CoT trajectories, enabling subsequent MCTS to operate over explicit reasoning paths rather than unstructured free-form responses.

\paragraph{Reasoning-conditioned Preference Optimization.}
As revealed in Section~\ref{sec:motivation}, response-level DPO behaves similarly to answer-only optimization, suggesting that reasoning chains may be under-emphasized when the CoT and final answer are jointly optimized as a complete response. To address this issue, we first formulate a CoT-oriented preference term that keeps the preferred answer fixed and compares different reasoning chains under the same multimodal input:
\begin{equation}
(c_w,a_w)|x \succ (c_l,a_w)|x,
\end{equation}
where \(x=(v,q)\), \(c_w\) denotes a visually grounded and logically valid CoT, and \(c_l\) denotes a hallucinated or corrupted CoT.

By the chain rule, the likelihood ratio of this preference can be decomposed as
\begin{equation}
\begin{aligned}
\log \frac{\pi_\theta(c_w,a_w|x)}
{\pi_\theta(c_l,a_w|x)}
&=
\log \frac{\pi_\theta(c_w|x)}
{\pi_\theta(c_l|x)} \\
&+
\log \frac{\pi_\theta(a_w|x,c_w)}
{\pi_\theta(a_w|x,c_l)}.
\end{aligned}
\end{equation}
The first term compares the likelihood of generating a high-quality CoT against a corrupted one, while the second term quantifies the conditional support that each CoT provides for generating the same preferred answer. 
Notably, the first term is essentially covered by standard response-level DPO over complete responses, whose likelihood ratio can be decomposed as
\begin{equation}
\begin{aligned}
\log \frac{\pi_\theta(c_w,a_w|x)}
{\pi_\theta(c_l,a_l|x)}
&=
\boxed{\displaystyle
\log \frac{\pi_\theta(c_w|x)}
{\pi_\theta(c_l|x)}
} \\
&+
\log \frac{\pi_\theta(a_w|x,c_w)}
{\pi_\theta(a_l|x,c_l)}.
\end{aligned}
\end{equation}

Hence, RC-DPO retains response-level DPO for general response preference learning and adds a reasoning-conditioned objective to explicitly optimize the answer-support term.
Following the DPO formulation, we define the reasoning-conditioned preference objective as
\begin{equation}
\begin{aligned}
\mathcal{L}_{\mathrm{RC}}
= -\log \sigma \Bigg(
&\beta \Bigg[
\log \frac{\pi_\theta(a_w|x,c_w)}
{\pi_{\mathrm{ref}}(a_w|x,c_w)} \\
&-
\log \frac{\pi_\theta(a_w|x,c_l)}
{\pi_{\mathrm{ref}}(a_w|x,c_l)}
\Bigg]
\Bigg),
\end{aligned}
\end{equation}
where \(\pi_{\mathrm{ref}}\) denotes the fixed reference model, instantiated as the SFT-warmed model \(\pi_{\mathrm{SFT}}\). By keeping the target answer fixed and varying only the CoT condition, this objective isolates the effect of the reasoning chain on answer generation. With the reference model as a baseline, the margin measures how much the policy model improves the conditional support for \(a_w\) under different CoT conditions. Maximizing this margin encourages the model to assign higher preference to the answer under a high-quality CoT than under a corrupted one, thereby providing a direct training signal for answer-supportive CoT alignment.

\paragraph{Training Objective.} 
The final RC-DPO objective combines response-level and reasoning-conditioned preference learning:
\begin{equation}
\mathcal{L}
=
\mathcal{L}_{\mathrm{DPO}}
+
\lambda
\mathcal{L}_{\mathrm{RC}},
\end{equation}
where \(\lambda\) controls the trade-off between the two objectives. The response-level objective maintains the general preference alignment over complete responses, while the reasoning-conditioned objective provides explicit supervision for CoT-level alignment by contrasting the conditional support for the preferred answer. This joint optimization enables RC-DPO to mitigate hallucinations in both intermediate reasoning chains and final answers.

\subsection{Preference Data Construction}

To support the above objective, RC-DPO requires reasoning-conditioned preference samples \((x,c_w, a_w, c_l,a_l)\). Unlike previous methods that rely on ground-truth annotations or response correction from stronger external models such as GPT \cite{yang2025mitigating}, we introduce a preference data generation pipeline based on the model's own outputs. 
Below, we detail the construction of positive and negative reasoning samples, respectively.


\paragraph{Positive Sample Construction.}
For each multimodal input \(x=(v,q)\), we use the SFT-warmed model $\pi_\mathrm{SFT}$ to perform reasoning search with Monte Carlo Tree Search (MCTS). Starting from the root node, MCTS incrementally expands candidate reasoning steps and performs rollouts to complete CoT-answer trajectories. During the search, a verifier is employed to evaluate candidate trajectories based on visual consistency, logical validity, and answer correctness. The resulting scores are back-propagated along the search tree to guide subsequent exploration. After the search, we select the highest-scoring trajectory as the positive sample:
\begin{equation}
(c_w,a_w)=\arg\max_{(c,a)\in \mathcal{T}(x)} S(c,a;x), 
\label{eq:mcts}
\end{equation}
where \(\mathcal{T}(x)\) denotes the set of candidate trajectories explored during the search, and \(S(\cdot)\) represents the verifier score. 
By selecting the highest-scoring trajectory, this process promotes positive CoTs with stronger visual grounding and logical consistency.
Additional details of the search algorithm are provided in Appendix \ref{appendix:mcts}.

\begin{table*}[htbp]
  \centering
  \caption{Performance comparison of RC-DPO with baseline methods on the Object HalBench. $C_S$ and $C_I$ denote CHAIR$_S$ and CHAIR$_I$ respectively, where lower values indicate fewer hallucinations.}
  \resizebox{0.8\linewidth}{!}{
    \begin{tabular}{l|cc|cc|cc|cc}
    \toprule
    \multicolumn{1}{l|}{\multirow{2}[0]{*}{Method}} & \multicolumn{2}{c|}{\textbf{R1-Onevision}} & \multicolumn{2}{c|}{\textbf{MM-Eureka}} & \multicolumn{2}{c|}{\textbf{ThinkLite-VL}} & \multicolumn{2}{c}{\textbf{OpenVLThinker}} \\
             & $C_S\downarrow$      & $C_I\downarrow$      & $C_S\downarrow$      & $C_I\downarrow$      & $C_S\downarrow$      & $C_I\downarrow$      & $C_S\downarrow$      & $C_I\downarrow$ \\
             \midrule
    \textit{Vanilla}  & 36.7     & 7.6      & 34.0     & 7.5      & 30.7     & 7.5      & 31.7     & 7.3 \\
    VCD      & 42.3     & 9.4      & 28.7     & 8.2      & 35.0     & 8.2      & 27.0     & 8.1 \\
    SID      & 43.7     & 9.2      & 35.3     & 8.3      & 34.7     & 8.4      & 35.7     & 8.9 \\
    RLAIF-V  & 38.0     & 7.9      & 33.3     & 7.3      & 33.0     & 7.8      & 27.3     & 6.8 \\
    OPA-DPO  & 40.0     & 8.6      & 29.0     & 5.4      & 30.3     & 7.6      & 27.0     & 7.1 \\
    \gc RC-DPO   & \gc\textbf{21.7}     & \gc\textbf{6.2}      & \gc\textbf{21.3}     & \gc\textbf{5.4}      &    \gc\textbf{18.7}      &   \gc\textbf{4.9}       &    \gc\textbf{24.3}      &  \gc\textbf{6.4}  \\
    \bottomrule
    \end{tabular}%
    }
  \label{tab:chair}%
\end{table*}%

\paragraph{Negative Sample Construction.}
To construct challenging negative CoTs, we introduce an image-attention-based CoT corruption strategy.
Given a response \((c,a)\) obtained by standard sampling from \(\pi_{\mathrm{SFT}}\), we compute the attention from each CoT token to image tokens. Specifically, for each CoT token \(c_t\), its image-attention score is computed as
\begin{equation}
s(c_t)
=
\frac{1}{|\mathcal{L}||\mathcal{H}|}
\sum_{l\in\mathcal{L}}
\sum_{h\in\mathcal{H}}
\sum_{k\in\mathcal{I}}
A^{(l,h)}_{t,k},
\end{equation}
where \(\mathcal{L}\) and \(\mathcal{H}\) denote the sets of selected attention layers and heads, respectively. \(\mathcal{I}\) denotes image-token positions, and \(A^{(l,h)}_{t,k}\) represents the attention weight from CoT token \(c_t\) to image token \(v_k\).

Intuitively, a higher image-attention score indicates that the token carries more visual information. We rank CoT tokens according to their scores \(s\) and remove the top-ranked visually salient tokens:
\begin{equation}
c_l = \mathrm{Prune}(c; s, r),
\label{eq:prune}
\end{equation}
where \(r\) denotes the pruning ratio. Since the pruned tokens are highly associated with visual evidence, the resulting CoT \(c_l\) disrupts the visual grounding of the reasoning chain, thereby serving as a challenging negative reasoning condition for RC-DPO.
\section{Experiments}

\subsection{Experimental Setup}
\paragraph{Models and Datasets.} We evaluate RC-DPO on four representative MLRMs: R1-Onevision-7B~\cite{yang2025r1}, MM-Eureka-7B~\cite{meng2025mm}, ThinkLite-VL-7B~\cite{wang2026sota}, and OpenVLThinker-7B~\cite{deng2026openvlthinker}. For training data construction, we adopt the RLAIF-V dataset~\cite{yu2025rlaif}, using 10K samples for SFT and 5K samples for DPO.

\paragraph{Evaluation Metrics.}
We comprehensively evaluate the performance across a wide range of benchmarks. For hallucination evaluation, we adopt Object HalBench~\cite{rohrbach2018object}, POPE~\cite{li2023evaluating}, MMHal-Bench~\cite{sun2024aligning}, GPT-4 assisted evaluation~\cite{zhao2023beyond}, and AMBER~\cite{wang2023amber}, covering object hallucination, response faithfulness, and hallucination-oriented multimodal reliability. For general-purpose evaluation, we use MME~\cite{fu2026mme}, MMBench~\cite{liu2024mmbench}, VMCBench~\cite{zhang2025automated}, and MMVP~\cite{tong2024eyes} to assess multimodal perception, reasoning, and visual-language understanding abilities. Detailed descriptions of these evaluation benchmarks are provided in Appendix~\ref{appendix:evaluation_benchmarks}.

\paragraph{Baseline algorithms.}
We compare RC-DPO with both decoding-based and training-based methods. For decoding-based baselines, we adopt VCD~\cite{leng2024mitigating} and SID~\cite{huo2025self}, which adjust token probability distributions through contrastive decoding. For training-based baselines, we adapt two state-of-the-art MLLM alignment methods, namely RLAIF-V~\cite{yu2025rlaif} and OPA-DPO~\cite{yang2025mitigating}.

\begin{table*}[htbp]
  \centering
  \caption{Comparison of RC-DPO with baseline methods on the POPE benchmark.}
  \resizebox{0.95\linewidth}{!}{
    \begin{tabular}{clcccccc}
    \toprule
    \multirow{2}[0]{*}{Model} & \multirow{2}[0]{*}{Method} & \multicolumn{2}{c}{\textbf{Random}} & \multicolumn{2}{c}{\textbf{Popular}} & \multicolumn{2}{c}{\textbf{Adversarial}} \\
    \cmidrule(lr){3-4} \cmidrule(lr){5-6} \cmidrule(lr){7-8}
             &          & Accuracy & F1 Score & Accuracy & F1 Score & Accuracy & F1 Score \\
             \midrule
    \multirow{6}[0]{*}{MM-Eureka} & \textit{Vanilla}  & 86.43\%  & 84.55\%  & 85.30\%  & 83.48\%  & 85.03\%  & 83.23\% \\
             & VCD      & 87.73\%  & 86.15\%  & 86.43\%  & 84.91\%  & 85.67\%  & 84.21\% \\
             & SID      & 87.50\%  & 85.88\%  & 85.93\%  & 84.35\%  & 85.40\%  & 83.85\% \\
             & RLAIF-V  & 86.30\%  & 84.28\%  & 84.87\%  & 82.92\%  & 84.50\%  & 82.57\% \\
             & OPA-DPO  & 87.33\%  & 85.63\%  & 86.13\%  & 84.49\%  & 85.30\%  & 83.71\% \\
             & \gc RC-DPO   & \gc\textbf{89.17\%} & \gc\textbf{87.95\%} & \gc\textbf{86.80\%} & \gc\textbf{85.69\%} & \gc\textbf{86.07\%} & \gc\textbf{85.02\%} \\
             \midrule
    \multirow{6}[0]{*}{R1-Onevision} & \textit{Vanilla}  & 76.83\%  & 76.46\%  & 75.93\%  & 75.76\%  & 76.47\%  & 76.16\% \\
             & VCD      & 76.57\%  & 76.90\%  & 76.27\%  & 76.38\%  & 76.80\%  & 76.97\% \\
             & SID      & 76.77\%  & 76.82\%  & 77.63\%  & 77.63\%  & 77.03\%  & 77.24\% \\
             & RLAIF-V  & 79.73\%  & 78.86\%  & 79.07\%  & 78.36\%  & 78.60\%  & 77.98\% \\
             & OPA-DPO  & 85.53\%  & 83.58\%  & 83.60\%  & 81.83\%  & 83.20\%  & 81.47\% \\
             & \gc RC-DPO   & \gc\textbf{87.27\%} & \gc\textbf{85.56\%} & \gc\textbf{85.50\%} & \gc\textbf{83.88\%} & \gc\textbf{84.80\%} & \gc\textbf{83.24\%} \\
             \bottomrule
    \end{tabular}%
    }
  \label{tab:pope}%
\end{table*}%

\begin{figure*}[t]
\begin{center}
\includegraphics[width=\linewidth]{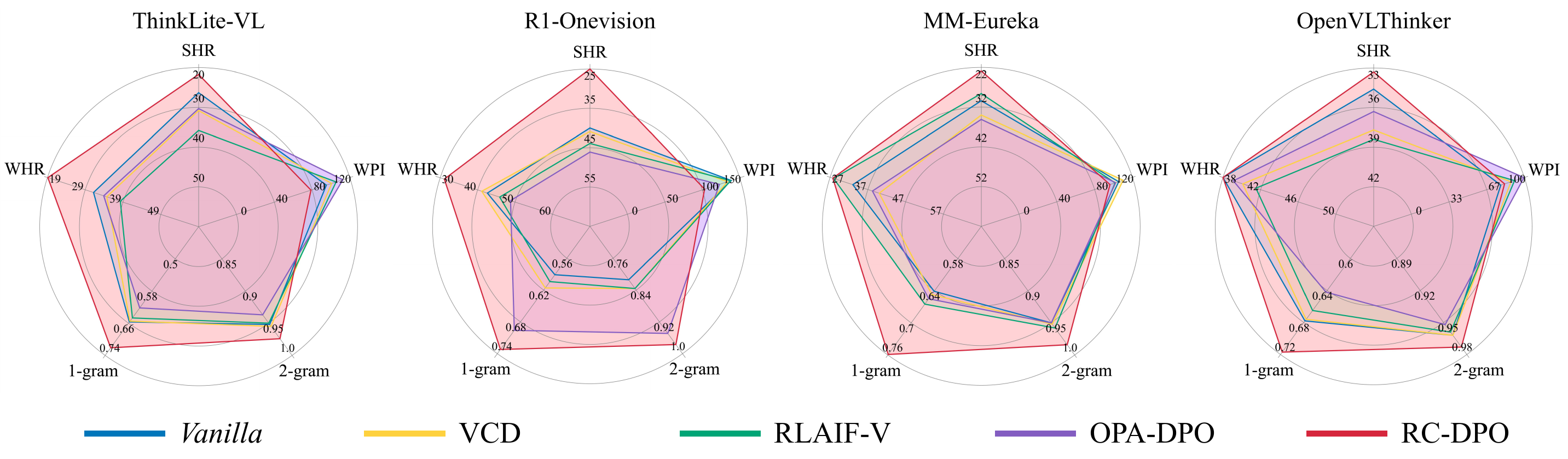}
\end{center}
\caption{Radar chart of the GPT-4 assisted evaluation results. We report the Sentence-level Hallucination Ratio (SHR), Word-level Hallucination Ratio (WHR), 1\&2-gram scores, and Words Per Image (WPI). A larger radar area indicates better overall performance.}
\label{fig:radar}
\end{figure*}

\paragraph{Implementation Details.}
In the warm-up stage, we perform SFT with Low-Rank Adaptation (LoRA), using a rank of 8 and training for 1 epoch with a learning rate of \(5e^{-5}\). 
For positive sample construction, we employ MCTS with a maximum of 3 children per node and 30 search iterations to discover high-quality responses. 
For negative sample construction, we use all attention layers and heads to compute image-attention scores, and set the pruning ratio \(r\) to 20\%. 
In the preference learning stage, we train the model for 1 epoch with a learning rate of \(1e^{-6}\) and a batch size of 32. The KL penalty coefficient is set to \(\beta=0.1\), and the reasoning-conditioned loss weight is set to \(\lambda=0.1\). More details are provided in Appendix \ref{appendix:implementation_details}.

\subsection{Performance Evaluation}
Due to space limitations, we report the results on Object HalBench, POPE, and GPT-4 assisted evaluation in this section. More comprehensive results are provided in Appendix~\ref{appendix:experimental_results}.

\paragraph{Object HalBench Evaluation.}
Following previous works~\cite{yu2025rlaif, yang2025mitigating}, we prompt MLRMs to generate captions for 300 images from the MSCOCO validation set~\cite{lin2014microsoft} and compute CHAIR metrics based on the generated outputs and object annotations. As shown in Table~\ref{tab:chair}, RC-DPO consistently achieves the lowest CHAIR$_S$ and CHAIR$_I$ scores across all evaluated MLRMs, demonstrating its effectiveness in reducing object-level hallucinations. Specifically, compared with the vanilla models, RC-DPO substantially reduces CHAIR$_S$ from 36.7 to 21.7 on R1-Onevision and from 34.0 to 21.3 on MM-Eureka. In contrast, decoding-based methods exhibit unstable performance and even increase hallucination in several cases, while training-based baselines yield limited and inconsistent improvements. These results indicate that RC-DPO more effectively mitigates object hallucinations in MLRMs by explicitly aligning reasoning chains.

\begin{figure*}[t]
\begin{center}
\includegraphics[width=\linewidth]{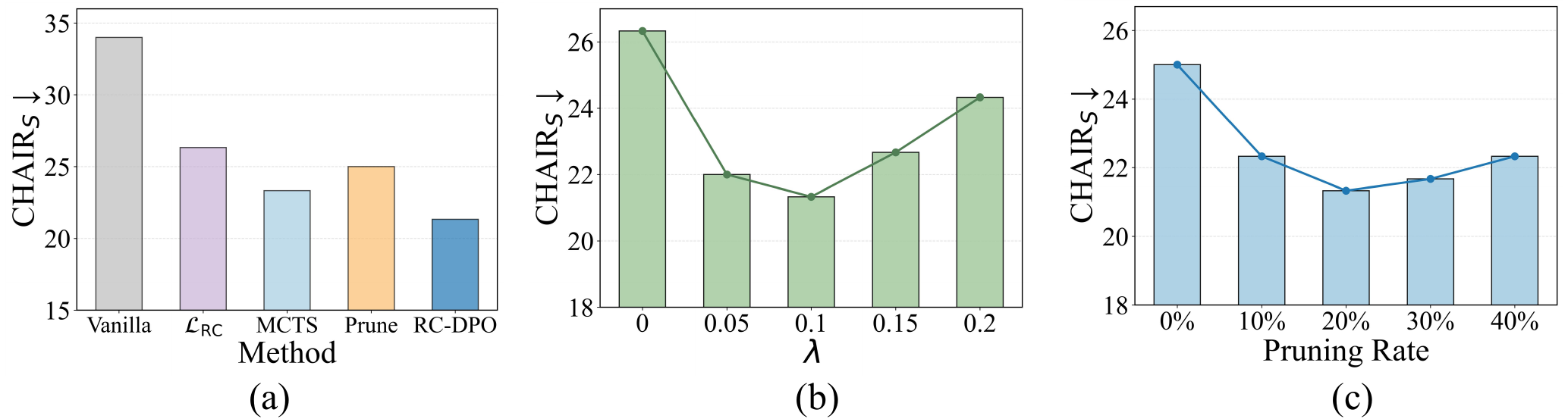}
\end{center}
\vspace{-0.5em}
\caption{Ablation studies of RC-DPO. (a) Effects of different components on CHAIR$_S$. (b) Sensitivity analysis of the reasoning-conditioned loss weight \(\lambda\). 
(c) Sensitivity analysis of the pruning rate for negative CoT construction.}
\label{fig:ablation}
\end{figure*}

\paragraph{POPE Evaluation.} 
The POPE benchmark evaluates object hallucinations with prompts in the format ``\texttt{Is there a <object> in the image?}''. The resulting yes/no responses are compared with ground-truth labels to compute accuracy, precision, recall, and F1 score.  
Due to space limitations, Table~\ref{tab:pope} reports results on MM-Eureka and R1-Onevision, while results on the other two models are provided in Appendix~\ref{appendix:pope}.
RC-DPO consistently achieves the best performance across all settings on both MM-Eureka and R1-Onevision. 
Compared with the strongest training-based baseline OPA-DPO, RC-DPO improves the F1 score by 2.32\%, 1.20\%, and 1.31\% on MM-Eureka under the three settings, respectively.
These results demonstrate that RC-DPO effectively improves object existence recognition and reduces hallucinated responses across diverse evaluation settings.

\paragraph{GPT-4 Assisted Evaluation.}
While POPE and CHAIR are reliable metrics widely adopted in previous studies, they mainly focus on object hallucination and may overlook other hallucination types. 
For a more comprehensive evaluation, we employ the GPT-4 assisted evaluation benchmark to assess both hallucination and descriptive quality in open-ended image captioning. Specifically, we report Sentence-level Hallucination Ratio (SHR), Word-level Hallucination Ratio (WHR), 1\&2-gram scores, and Words Per Image (WPI) in Figure~\ref{fig:radar}. RC-DPO achieves consistently strong performance across four MLRMs. Compared with vanilla models and baseline methods, RC-DPO generally covers a larger radar area, indicating better overall performance. The improvements on SHR and WHR demonstrate its effectiveness in reducing hallucinated descriptions, while the high 1\&2-gram scores indicate that RC-DPO generates adequate and fluent responses. These results further verify that reasoning-chain alignment improves response faithfulness without sacrificing descriptive quality.


\subsection{Ablation Study}

\paragraph{Module ablation.}
We evaluate the contribution of each component in RC-DPO. As shown in Figure~\ref{fig:ablation} (a), removing the reasoning-conditioned loss \(\mathcal{L}_{\mathrm{RC}}\), MCTS-based sampling, or attention-guided pruning consistently increases CHAIR$_S$. Among them, removing \(\mathcal{L}_{\mathrm{RC}}\) causes the most significant performance drop, highlighting the importance of explicitly optimizing answer-supportive CoT alignment. The full RC-DPO achieves the best performance, demonstrating that positive reasoning search and negative CoT corruption provide complementary benefits for hallucination mitigation.

\paragraph{Hyperparameter ablation.}
We further study the sensitivity of RC-DPO to the loss weight \(\lambda\) and pruning rate \(r\). As shown in Figure~\ref{fig:ablation} (b), increasing \(\lambda\) initially improves performance, with the best result achieved at \(\lambda=0.1\). However, larger values lead to performance degradation, suggesting that excessive RC supervision may hinder general preference learning. 
Figure~\ref{fig:ablation} (c) shows that a pruning rate of 20\% achieves the lowest CHAIR$_S$. A smaller pruning rate may produce weak negative CoTs, while a larger rate may overly disrupt the reasoning structure. Therefore, we set \(\lambda=0.1\) and \(r=20\%\) in our main experiments.




\section{Conclusion}

In this paper, we investigate hallucination mitigation in MLRMs from the perspective of reasoning-chain alignment. Motivated by our observation that response-level DPO primarily captures answer-level preference while leaving reasoning chains insufficiently optimized, we propose RC-DPO, a reasoning-conditioned preference optimization framework that explicitly treats the CoT as a condition for answer generation and aligns reasoning chains by comparing their support for the preferred answer. 
To enable this objective, we further introduce a preference data construction pipeline that combines MCTS-based positive reasoning search with image-attention-guided CoT corruption for negative sample construction. Extensive experiments across multiple MLRMs and benchmarks demonstrate that RC-DPO effectively reduces hallucinations while improving general multimodal reasoning ability. These results highlight the importance of explicitly aligning reasoning chains for building more faithful and reliable MLRMs.

\section*{Ethical Statement}
This work aims to improve the faithfulness and reliability of Multimodal Large Reasoning Models by mitigating hallucinations in both reasoning chains and final answers. We conduct experiments on publicly available datasets and benchmarks, and do not collect or annotate any private user data. We adhere to strict ethical standards throughout our study. All code, models, and datasets used in this study are employed in accordance with their intended use and corresponding licenses.
Since our method is designed to reduce hallucinated content rather than generate harmful or sensitive outputs, we do not intentionally introduce unsafe data or malicious instructions during training. We also report results across diverse benchmarks to avoid over-claiming effectiveness on a single evaluation setting.

\section*{Limitation}
Although RC-DPO demonstrates strong effectiveness in mitigating hallucinations in MLRMs, there remain several directions for future improvement. 
\begin{itemize}
    \item Our work makes an initial step toward reasoning-chain alignment for hallucination mitigation in MLRMs at the trajectory level, where each CoT is treated as a complete reasoning condition for answer generation. Future work could explore more fine-grained reasoning-chain alignment, such as step-level preference modeling or token-level process supervision, to better understand which intermediate reasoning steps contribute most to faithful answer generation.
    \item Our current negative sample construction focuses on corrupting CoTs by pruning visually salient tokens based on image attention. Future work could further enrich the preference signal by incorporating more diverse reasoning conditions, such as subtle logical inconsistencies, commonsense conflicts, or multi-step reasoning failures, enabling more comprehensive reasoning-chain alignment.
\end{itemize}

\bibliography{custom}

\appendix
\clearpage

\section{Pseudocode}
\label{appendix:pseudocode}

The pseudocode of our RC-DPO is provided in Alg.~\ref{alg:rcdpo}, with the definitions of the associated loss functions detailed in the main text.

\begin{algorithm}[]
    \caption{Pseudocode of RC-DPO}
    \label{alg:rcdpo}
    \begin{algorithmic}[1]
        \Require 
        $\pi_0$: the base MLRM;
        $\mathcal{D}$: the image-question dataset;
        $S(\cdot)$: the verifier model;
        $r$: the pruning ratio;
        $N_{\mathrm{sft}}$: the number of SFT iterations;
        $N_{\mathrm{dpo}}$: the number of DPO iterations;
        \Ensure 
        optimized policy model $\pi_\theta$;
        \State Initialize $\pi_\theta \leftarrow \pi_0, \mathcal{D}_{\mathrm{SFT}} \leftarrow \emptyset, \mathcal{D}_{\mathrm{pref}} \leftarrow \emptyset$;
        \State \texttt{// CoT Format Alignment}
        \For{each input $x=(v,q) \in \mathcal{D}$}
            \State Prompt $\pi_\theta$ to generate step-wise $c$ and $a$;
            \State Add $(x,c,a)$ to $\mathcal{D}_{\mathrm{SFT}}$;
        \EndFor
        \For{$i \leftarrow 1$ to $N_{\mathrm{sft}}$}
            \State Update $\pi_\theta$ on $\mathcal{D}_{\mathrm{SFT}}$ using $\mathcal{L}_{\mathrm{SFT}}$ in Eq. \ref{eq:sft};
        \EndFor
        \State Set the SFT-warmed model $\pi_{\mathrm{SFT}} \leftarrow \pi_\theta$;

        \State \texttt{// Preference Data Construction}
        \For{each input $x=(v,q) \in \mathcal{D}$}
            \State Conduct MCTS with $\pi_{\mathrm{SFT}}$ and $S(\cdot)$;
            \State Select $(c_w,a_w)$ using Eq.~\ref{eq:mcts};
            \State Sample $(c,a) \sim \pi_{\mathrm{SFT}}(\cdot|x)$;
            \State Compute $s(c_t)$ for each CoT token $c_t$;
            \State $c_l \leftarrow \mathrm{Prune}(c;s,r)$ by Eq.~\ref{eq:prune};
            \State Add $(x,c_w,a_w,c_l,a)$ to $\mathcal{D}_{\mathrm{pref}}$;
        \EndFor

        \State \texttt{// RC-DPO Training}
        \State Initialize reference model $\pi_{\mathrm{ref}} \leftarrow \pi_{\mathrm{SFT}}$;
        \For{$i \leftarrow 1$ to $N_{\mathrm{dpo}}$}
            \State Sample a mini-batch from $\mathcal{D}_{\mathrm{pref}}$;
            \State Compute response-level loss $\mathcal{L}_{\mathrm{DPO}}$;
            \State Compute reasoning-conditioned loss $\mathcal{L}_{\mathrm{RC}}$;
            \State Update $\pi_\theta$ using $\mathcal{L}=\mathcal{L}_{\mathrm{DPO}}+\lambda\mathcal{L}_{\mathrm{RC}}$;
        \EndFor

        \State \textbf{return} optimized policy model $\pi_\theta$;
    \end{algorithmic}
\end{algorithm}

\section{Related Work}

\subsection{Multimodal Large Reasoning Models}

Large Reasoning Models have recently shown strong capabilities on complex tasks by explicitly generating intermediate inference steps before producing final responses \cite{wang2022self, guo2025deepseek}. Inspired by their success, recent studies have extended this reasoning-oriented paradigm to vision-language scenarios, giving rise to Multimodal Large Reasoning Models \cite{leng2025mmr1, zhan2025vision}. Early multimodal reasoning works construct step-by-step rationales for visual question answering and science reasoning \cite{lu2022learn, zhang2023multimodal}, while more advanced MLRMs enhance long-chain visual reasoning through supervised fine-tuning (SFT) \cite{yang2025r1} and reinforcement learning (RL) \cite{ma2025one}. Various techniques, including data construction \cite{meng2025mm}, sample selection \cite{wang2026sota}, and self-improvement \cite{deng2026openvlthinker} have also been explored to further elicit and strengthen multimodal reasoning abilities. 

\subsection{Multimodal Hallucination}

Extensive studies have demonstrated that multimodal models are prone to hallucination, \textit{i.e.}, generating fluent and plausible responses that are inconsistent with the visual content~\cite{liu2024survey, bai2024hallucination, yang2026looking}. Existing hallucination mitigation methods for MLLMs can be broadly categorized into training-based and training-free approaches. Training-based methods mainly improve cross-modal alignment through high-quality instruction data or advanced alignment algorithms~\cite{liu2024mitigating, yu2024rlhf}. In contrast, training-free methods intervene during inference stage, including contrastive decoding~\cite{leng2024mitigating, wang2024mitigating}, token pruning~\cite{huo2025self, zhuang2025vasparse}, latent steering~\cite{liu2025reducing}, and mutual-information-based calibration~\cite{fang2026grounding}.

In MLRMs, hallucination becomes more subtle and challenging, since extended reasoning may amplify reliance on language priors and reduce attention to visual evidence \cite{xu2026more,fang2026seeing}. Recent work has investigated the relationship between reasoning ability and hallucination severity \cite{xu2026more}, while another line analyzes hallucinations within multimodal reasoning chains and introduces benchmarks tailored to reasoning hallucination evaluation \cite{dong2026mirage}. 
To mitigate hallucinations in MLRMs, recent studies have explored post-alignment with curated comparison pairs, including AI-feedback-based alignment \cite{yu2025rlaif}, on-policy optimization \cite{yang2025mitigating}, and CoT compression \cite{fang2026seeing}. However, these methods typically rely on coarse-grained supervision, leaving reasoning chains insufficiently optimized.

\section{MCTS for Positive Data Construction}
\label{appendix:mcts}
To construct high-quality positive reasoning samples, we employ Monte Carlo Tree Search (MCTS) over structured CoT trajectories. For each multimodal input \(x=(v,q)\), the SFT-warmed model \(\pi_{\mathrm{SFT}}\) serves as the policy model for expanding reasoning paths, while a verifier \(S(\cdot)\) evaluates the quality of completed CoT-answer trajectories. Each node in the search tree corresponds to a partial reasoning trajectory, and each edge represents a newly generated reasoning step. The goal of MCTS is to explore diverse reasoning paths and identify a trajectory that is visually grounded, logically consistent, and leads to a correct final answer.
Specifically, MCTS iteratively performs four steps: selection, expansion, simulation, and backpropagation.

\subsection{Selection}
Starting from the root node corresponding to the input \(x=(v,q)\), MCTS selects a promising path from the current search tree. Following the standard MCTS procedure~\cite{vodopivec2017monte}, we use UCB1 algorithm~\cite{chang2005adaptive} to select the child node \(s'\) at each intermediate node \(s\):
\begin{equation}
s^*
=
\arg\max_{s' \in \mathrm{Child}(s)}
\left[
Q(s')
+
\alpha
\sqrt{
\frac{\log N(s)}
{N(s')+\epsilon}
}
\right],
\end{equation}
where \(Q(s')\) denotes the estimated value of child node \(s'\), \(N(s)\) and \(N(s')\) are the visit counts of the parent and child nodes, respectively, \(\alpha\) controls the exploration strength, and \(\epsilon\) is a small constant for numerical stability. The first term encourages exploitation of high-value reasoning paths, while the second term promotes exploration of less-visited nodes, thereby balancing trajectory quality and diversity. This selection process continues until reaching a leaf node that has not been expanded.

\subsection{Expansion.}
After the selection stage reaches an expandable leaf node \(s\), MCTS expands the search tree by generating new child nodes from \(s\). In our setting, each node represents a partial reasoning trajectory, and each child node corresponds to extending the current trajectory with a newly generated reasoning step. Specifically, we use the SFT-warmed model \(\pi_{\mathrm{SFT}}\) to sample candidate next steps conditioned on the input \(x\) and the current partial CoT \(c_{1:t}\):
\begin{equation}
c_{t+1} \sim \pi_{\mathrm{SFT}}(\cdot|x,c_{1:t}).
\end{equation}
Each sampled step is appended to the current trajectory to form a new child node. To control the search cost, we restrict the maximum number of children for each node. In this way, MCTS explores multiple possible reasoning directions while maintaining a manageable search space.

\subsection{Simulation.}
For each newly expanded node, MCTS performs a rollout to estimate the quality of the corresponding reasoning path. Starting from the partial trajectory \(c_{1:t+1}\), we use \(\pi_{\mathrm{SFT}}\) to complete the remaining reasoning process and generate a full CoT-answer trajectory \((c,a)\). The completed trajectory is then evaluated by the verifier \(S(\cdot)\):
\begin{equation}
R = S(c,a;x),
\end{equation}
where \(R\) denotes the reward assigned to the rollout trajectory. The verifier considers the visual consistency of the CoT with the input image, the logical coherence of the reasoning process, and the correctness of the final answer. Therefore, trajectories that are visually grounded, logically valid, and answer-correct receive higher rewards, which are later used to guide the tree search.

\subsection{Backpropagation.}
After obtaining the verifier reward \(R\), MCTS back-propagates it along the selected path from the expanded node to the root. For each visited node \(s\), we update its visit count and value estimate:
\begin{equation}
N(s) \leftarrow N(s)+1,
\end{equation}
\begin{equation}
Q(s) \leftarrow Q(s)+\frac{R-Q(s)}{N(s)}.
\end{equation}
Here, \(N(s)\) records how many times node \(s\) has been visited, and \(Q(s)\) denotes the average reward of trajectories passing through \(s\). This update allows high-quality reasoning paths to receive larger value estimates, making them more likely to be selected in subsequent iterations. 

By repeatedly performing selection, expansion, simulation, and backpropagation, MCTS gradually concentrates the search on promising reasoning trajectories. After the search terminates, we select the trajectory with the highest verifier score as the positive CoT-answer sample:
\[
(c_w,a_w)=\arg\max_{(c,a)\in \mathcal{T}(x)} S(c,a;x),
\]
where \(\mathcal{T}(x)\) denotes the set of completed trajectories explored during MCTS. The selected trajectory is then used as the positive CoT-answer pair for subsequent preference optimization.

\section{Experimental Details}
\label{appendix:experimental_details}

\subsection{Evaluated Models}
As mentioned in the main text, we adopt R1-Onevision-7B~\cite{yang2025r1}, MM-Eureka-7B~\cite{meng2025mm}, ThinkLite-VL-7B~\cite{wang2026sota}, and OpenVLThinker-7B~\cite{deng2026openvlthinker}. 
These models are selected because they follow the reason-before-answer paradigm and generate explicit CoT-style reasoning before final answers, making them suitable for evaluating CoT-level hallucination mitigation.
Specifically, R1-Onevision is designed for general multimodal reasoning, while MM-Eureka and ThinkLite-VL focus on long-chain visual reasoning and multimodal problem solving. OpenVLThinker further enhances complex vision-language reasoning through iterative training. Their diverse training recipes and reasoning behaviors allow us to assess the generalizability of RC-DPO across different MLRMs.

\subsection{Evaluation Benchmarks}
\label{appendix:evaluation_benchmarks}

\paragraph{Object HalBench.}
Caption Hallucination Assessment with Image Relevance (CHAIR)~\cite{rohrbach2018object} is a widely used metric for evaluating object hallucination in image captioning. It measures whether generated captions mention objects that are not present in the ground-truth object annotations. 
Following prior work, we report both CHAIR$_S$ and CHAIR$_I$, defined as:
\[
C_I =
\frac{|\text{hallucinated objects}|}
{|\text{all mentioned objects}|},
\]
\[
C_S =
\frac{|\text{captions with hallucinated objects}|}
{|\text{all captions}|}.
\]
Here, \(C_S\) measures the proportion of captions containing at least one hallucinated object, while \(C_I\) measures the proportion of hallucinated object instances among all mentioned objects. Lower values indicate fewer object hallucinations.

\paragraph{POPE.}
Polling-based Object Probing Evaluation (POPE)~\cite{li2023evaluating} evaluates object hallucination through binary object-existence questions. Given an image, POPE asks questions in the format of ``\texttt{Is there a <object> in the image?}'' and compares the model's yes/no responses with ground-truth labels. It includes three evaluation settings: \textit{random}, \textit{popular}, and \textit{adversarial}, where absent objects are sampled randomly, from frequent object categories, or from objects that frequently co-occur with present objects. We report accuracy and F1 score, where higher values indicate stronger object grounding ability.

\paragraph{MMHal-Bench.}
MMHal-Bench~\cite{sun2024aligning} is designed to evaluate hallucinations in open-ended multimodal responses. It contains challenging image-question pairs that require models to generate visually grounded answers rather than relying on language priors. The benchmark evaluates both response quality and hallucination severity, making it suitable for assessing whether a model can produce helpful yet faithful answers. We use MMHal-Bench to examine whether RC-DPO reduces hallucinations in more natural open-ended vision-language interactions.

\begin{table*}[t]
  \centering
  \caption{Comparison of RC-DPO with baseline methods on the POPE benchmark.}
  \resizebox{0.95\linewidth}{!}{
    \begin{tabular}{clcccccc}
    \toprule
    \multirow{2}[0]{*}{Model} & \multirow{2}[0]{*}{Method} & \multicolumn{2}{c}{\textbf{Random}} & \multicolumn{2}{c}{\textbf{Popular}} & \multicolumn{2}{c}{\textbf{Adversarial}} \\
    \cmidrule(lr){3-4} \cmidrule(lr){5-6} \cmidrule(lr){7-8}
             &          & Accuracy & F1 Score & Accuracy & F1 Score & Accuracy & F1 Score \\
             \midrule
    \multirow{6}[0]{*}{ThinkLite-VL} & \textit{Vanilla}  & 88.37\%  & 86.98\%  & 86.73\%  & 85.41\%  & 85.73\%  & 84.48\% \\
             & VCD      & 88.57\%  & 87.23\%  & 86.97\%  & 85.74\%  & 86.07\%  & 84.87\% \\
             & SID      & 88.87\%  & 87.66\%  & 86.97\%  & 85.80\%  & 86.10\%  & 85.06\% \\
             & RLAIF-V  & 87.90\%  & 86.34\%  & 86.73\%  & 85.22\%  & 85.40\%  & 83.97\% \\
             & OPA-DPO  & 89.23\%  & 88.09\%  & 87.40\%  & 86.33\%  & \textbf{86.13\%} & 85.16\% \\
             & \gc RC-DPO   & \gc\textbf{90.73\%} & \gc\textbf{90.07\%} & \gc\textbf{88.03\%} & \gc\textbf{87.55\%} & \gc86.00\%  & \gc\textbf{85.73\%} \\
             \midrule
    \multirow{6}[0]{*}{OpenVLThinker} & \textit{Vanilla}  & 85.77\%  & 83.52\%  & 84.73\%  & 82.52\%  & 84.20\%  & 82.02\% \\
             & VCD      & 86.40\%  & 84.38\%  & 85.47\%  & 83.51\%  & 84.37\%  & 82.41\% \\
             & SID      & 86.77\%  & 84.86\%  & 85.73\%  & 83.90\%  & 84.53\%  & 82.63\% \\
             & RLAIF-V  & 84.53\%  & 81.78\%  & 83.63\%  & 80.86\%  & 82.93\%  & 80.20\% \\
             & OPA-DPO  & 87.27\%  & 85.53\%  & 85.87\%  & 84.19\%  & \textbf{85.20\%} & \textbf{83.57\%} \\
             & \gc RC-DPO   & \gc\textbf{87.63\%} & \gc\textbf{86.20\%} & \gc\textbf{86.00\%} & \gc\textbf{84.46\%} & \gc84.87\%  & \gc83.41\% \\
             \bottomrule
    \end{tabular}%
    }
  \label{tab:appendix_pope}%
\end{table*}%

\paragraph{GPT-4 Assisted Evaluation.} 
To quantify hallucinations in detailed image descriptions, we adopt GPT-4 assisted evaluation following established protocols. Specifically, models are prompted to generate detailed captions for images from the Visual Genome dataset~\cite{krishna2017visual}, and GPT-4 is used to identify hallucinated or contradictory statements by comparing the generated descriptions with the ground-truth annotations. This evaluation complements object-centric metrics by capturing broader hallucination types, such as attribute, position, and relation errors. Following the judge-based protocol, GPT-4 is instructed to mark statements that are inconsistent with the visual annotations and to summarize the degree of hallucination in the generated response.

\paragraph{AMBER.}
AMBER~\cite{wang2023amber} is an LLM-free multi-dimensional benchmark for evaluating hallucinations in MLLMs. It covers both generative and discriminative evaluation settings and considers multiple hallucination types, including existence, attribute, and relation hallucinations. Instead of relying on an external LLM judge, AMBER provides fine-grained annotations and automatic evaluation protocols, enabling scalable and reproducible hallucination assessment. 

\paragraph{MME.}
MME~\cite{fu2026mme} is a comprehensive benchmark for evaluating multimodal perception and cognition abilities. It contains fine-grained image-grounded questions across multiple categories, including object recognition, OCR, counting, commonsense reasoning, and other visual understanding tasks. The benchmark uses a unified question-answering format to assess whether models can produce accurate and image-grounded responses. It reports separate scores for perception and cognition, enabling a more detailed analysis of model capabilities. We use MME to examine whether RC-DPO preserves general multimodal understanding while mitigating hallucinations.

\begin{table*}[htbp]
  \centering
  \caption{Comparison of RC-DPO with baseline methods on the MMHal-Bench.}
  \resizebox{\linewidth}{!}{
    \begin{tabular}{lcccccccc}
    \toprule
    \multirow{2}[0]{*}{Method} & \multicolumn{2}{c}{MM-Eureka} & \multicolumn{2}{c}{R1-Onevision} & \multicolumn{2}{c}{ThinkLite-VL} & \multicolumn{2}{c}{OpenVLThinker} \\
    \cmidrule(lr){2-3} \cmidrule(lr){4-5} \cmidrule(lr){6-7}  \cmidrule(lr){8-9} 
             & Avg Score$\uparrow$ & Hal Rate$\downarrow$ & Avg Score$\uparrow$ & Hal Rate$\downarrow$ & Avg Score$\uparrow$ & Hal Rate$\downarrow$ & Avg Score$\uparrow$ & Hal Rate$\downarrow$ \\
             \midrule
    \textit{Vanilla}  & 3.33     & 54.17\%  & 2.66     & 65.62\%  & 2.56     & 68.75\%  & 2.91     & 63.54\% \\
    VCD      & 3.24     & 54.17\%  & 2.57     & 68.75\%  & 3.29     & 55.21\%  & 2.98     & 61.46\% \\
    SID      & 2.99     & 61.46\%  & 2.68     & 66.67\%  & 3.27     & 55.21\%  & 2.91     & 63.54\% \\
    RLAIF-V  & 3.10     & 57.29\%  & 2.97     & 59.38\%  & 3.01     & 60.42\%  & 2.83     & 61.46\% \\
    OPA-DPO  & 3.27     & 54.17\%  & 2.86     & 62.50\%  & 3.35     & 55.21\%  & 2.79     & 66.67\% \\
    \gc RC-DPO   & \gc\textbf{4.01} & \gc\textbf{40.62\%} & \gc\textbf{3.29} & \gc\textbf{55.21\%} & \gc\textbf{3.75} & \gc\textbf{43.75\%} & \gc\textbf{3.34} & \gc\textbf{52.08\%} \\
    \bottomrule
    \end{tabular}%
    }
  \label{tab:mmhal}%
\end{table*}%

\begin{figure*}[t]
\begin{center}
\includegraphics[width=\linewidth]{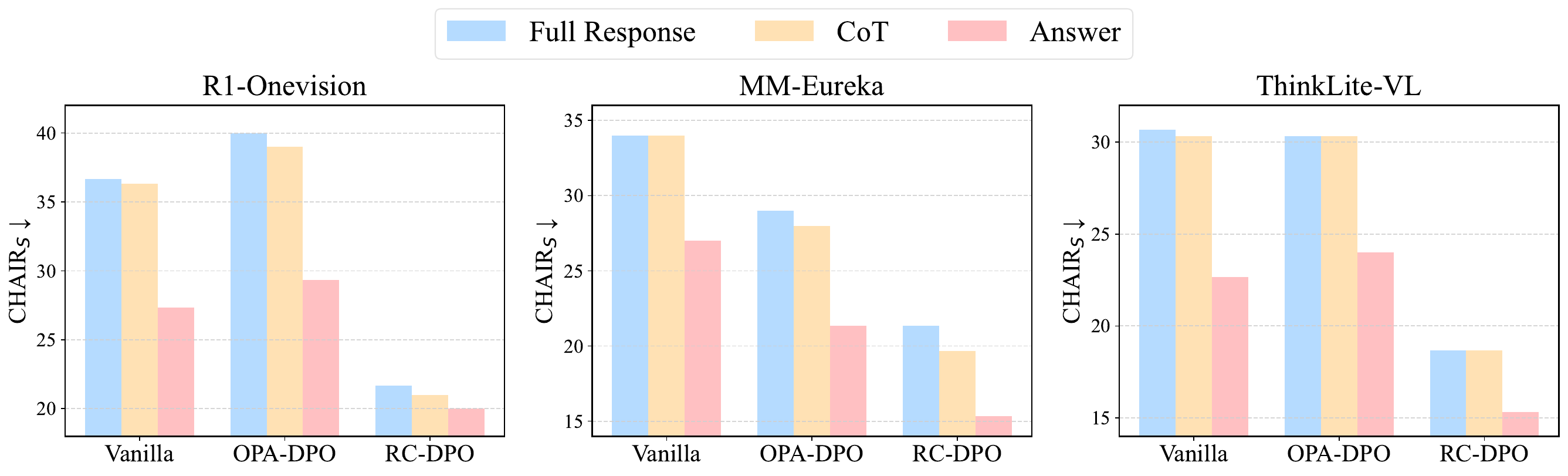}
\end{center}
\vspace{-0.5em}
\caption{Segment-level CHAIR$_S$ results on Object HalBench. We compare Vanilla, OPA-DPO, and RC-DPO across three MLRMs, and report CHAIR$_S$ scores for the full response, CoT, and final answer separately.}
\label{fig:chair}
\end{figure*}

\paragraph{MMBench.}
MMBench~\cite{liu2024mmbench} is a large-scale multimodal benchmark designed to evaluate diverse vision-language abilities. It contains carefully curated multiple-choice questions covering fine-grained perception, attribute recognition, spatial understanding, logical reasoning, and commonsense reasoning. MMBench also adopts robust evaluation strategies to reduce answer-position bias and improve evaluation reliability. 

\paragraph{VMCBench.}
VMCBench~\cite{zhang2025automated} is a multiple-choice vision-language benchmark constructed by converting diverse VQA datasets into a unified evaluation format. It covers a wide range of visual and linguistic contexts, including perception, recognition, reasoning, and commonsense understanding, and is designed to provide scalable, consistent, and reproducible evaluation of VLMs. By using multiple-choice questions, VMCBench reduces ambiguity in open-ended evaluation while preserving the difficulty of visual understanding tasks. Its diverse data sources and unified format make it suitable for assessing whether a method generalizes beyond a specific task or dataset. 
We use VMCBench to evaluate the generalization ability across diverse multimodal reasoning scenarios.

\paragraph{MMVP.}
MMVP~\cite{tong2024eyes} is designed to diagnose the visual shortcomings of multimodal models. It highlights failure cases where models tend to rely on language priors or insufficient visual representations, particularly when images contain subtle but semantically critical visual differences. The benchmark consists of challenging visual patterns that require careful perception and fine-grained discrimination. We use MMVP to assess whether fine-grained visual grounding ability is preserved after optimization.

\subsection{Implementation Details}
\label{appendix:implementation_details}

\paragraph{SFT Format Alignment.}
We conduct the SFT warm-up with LoRA for parameter-efficient fine-tuning. For all models, LoRA is applied to the language-model components, while the vision encoder remains frozen. The LoRA rank is set to 8, and the scaling factor is set to 16. We train each model for 1 epoch with a learning rate of \(5e^{-5}\), using AdamW as the optimizer and a linear warmup-decay scheduler. The maximum sequence length is set to 2048, and the global batch size is set to 64. Training is performed with bfloat16 precision and takes about 1.5 hours on four L20 GPUs.

\paragraph{Positive Sample Construction.}
For MCTS-based positive sample construction, we adopt the UCB1 algorithm for node selection to balance trajectory quality and diversity. To control the search cost, we set the maximum number of children per node to 3, the maximum depth to 10, and the maximum number of search iterations to 30. During the simulation stage, Qwen3-VL~\cite{bai2025qwen3} is used as the verifier to score each trajectory based on the prompt shown in Figure~\ref{fig:eval_prompt}.

\paragraph{Negative Sample Construction.}
For negative sample construction, we compute image-to-CoT attention scores to quantify the visual relevance of CoT tokens, and prune the top-ranked important tokens to construct corrupted CoTs. We use all attention layers and heads to calculate the scores and set the pruning ratio to \(r=20\%\). Since one word may correspond to multiple subword tokens, we apply max pooling over subword-level scores to obtain word-level importance scores and determine the words to be removed.

\paragraph{Preference learning.}
In the preference learning stage, we optimize the policy model with the combined RC-DPO objective. The SFT-warmed model \(\pi_{\mathrm{SFT}}\) is used as the fixed reference model \(\pi_{\mathrm{ref}}\). We train the model for 1 epoch with a learning rate of \(1e^{-6}\) and a batch size of 32. The maximum sequence length is set to 2048, and Adam is used as the optimizer. This training stage takes around 4 hours on four L20 GPUs.

\begin{table*}[htbp]
  \centering
  \caption{Comparison of RC-DPO with baseline methods on the AMBER benchmark.}
  \resizebox{\linewidth}{!}{
    \begin{tabular}{clcccccccc}
    \toprule
    \multirow{2}[0]{*}{Model} & \multirow{2}[0]{*}{Method} & \multicolumn{4}{c}{\textbf{Generative}}   & \multicolumn{4}{c}{\textbf{Discriminative}} \\
    \cmidrule(lr){3-6}
    \cmidrule(lr){7-10}
             &          & CHAIR$\downarrow$   & Cover$\uparrow$   & HalRate$\downarrow$ & Cog$\downarrow$     & Accuracy & Precision & Recall   & F1 Score \\
             \midrule
    \multirow{6}[0]{*}{MM-Eureka} & \textit{Vanilla} & 4.0      & 59.3     & 21.2     & 1.2      & 85.99    & 87.30    & 91.65    & 89.41 \\
             & VCD      & 4.8      & \textbf{61.5} & 24.3     & 1.4      & 86.22    & 87.60    & 91.63    & 89.57 \\
             & SID      & 4.6      & 60.7     & 23.7     & 1.1      & 86.02    & 87.61    & 91.13    & 89.33 \\
             & RLAIF-V  & 4.1      & 56.0     & 18.0     & 1.1      & 85.62    & 86.44    & \textbf{92.21} & 89.22 \\
             & OPA-DPO  & 5.1      & 57.5     & 21.5     & 1.6      & 86.23    & 87.28    & 92.11    & 89.63 \\
             & \gc RC-DPO   & \gc\textbf{3.9} & \gc56.2     & \gc\textbf{16.2} & \gc\textbf{1.0} & \gc\textbf{86.34} & \gc\textbf{87.62} & \gc91.83    & \gc\textbf{89.67} \\
             \midrule
    \multirow{6}[0]{*}{R1-Onevision} & \textit{Vanilla} & 6.6      & 63.7     & 35.6     & 2.1      & 70.23    & 85.28    & 65.16    & 73.87 \\
             & VCD      & 7.8      & 63.9     & 40.4     & 2.0      & 70.69    & 86.03    & 65.21    & 74.18 \\
             & SID      & 7.9      & \textbf{64.7} & 39.1     & 2.1      & 70.39    & 85.46    & 65.26    & 74.00 \\
             & RLAIF-V  & 6.8      & 64.4     & 36.8     & 2.4      & 72.40    & 86.09    & 68.32    & 76.18 \\
             & OPA-DPO  & 6.6      & 62.9     & 32.8     & 1.8      & 77.36    & 86.76    & 76.64    & 81.38 \\
             & \gc RC-DPO   & \gc\textbf{4.4} & \gc59.5     & \gc\textbf{20.8} & \gc\textbf{1.2} & \gc\textbf{85.28} & \gc\textbf{87.52} & \gc\textbf{90.05} & \gc\textbf{88.76} \\
             \midrule
    \multirow{6}[0]{*}{ThinkLite-VL} & \textit{Vanilla} & 4.0      & 58.3     & 18.3     & 1.0      & 86.38    & 88.97    & 90.09 & 89.52 \\
             & VCD      & 4.6      & 58.1     & 20.2     & 1.0      & 86.46    & \textbf{89.60} & 89.41    & 89.50 \\
             & SID      & 4.4      & 57.8     & 19.8     & 1.0      & 85.78    & 89.40    & 88.49    & 88.94 \\
             & RLAIF-V  & 4.4      & 57.9     & 21.6     & 1.5      & 86.42    & 89.13    & 89.95    & 89.53 \\
             & OPA-DPO  & 4.6      & \textbf{61.5} & 25.2     & 1.8      & 86.73    & 89.10    & \textbf{90.54}    & 89.81 \\
             & \gc RC-DPO   & \gc\textbf{3.8} & \gc56.8     & \gc\textbf{15.2} & \gc\textbf{1.0} & \gc\textbf{86.80} & \gc89.56    & \gc90.07    & \gc\textbf{89.81} \\
             \bottomrule
    \end{tabular}%
    }
  \label{tab:amber}%
\end{table*}%

\section{More Experimental Results}
\label{appendix:experimental_results}

\subsection{POPE Evaluation}
\label{appendix:pope}
Table~\ref{tab:appendix_pope} provides additional POPE results on ThinkLite-VL and OpenVLThinker, complementing the main results reported in Table~\ref{tab:pope}. Despite the stronger vanilla performance of these models, RC-DPO consistently improves over the vanilla baselines across all three POPE settings and achieves the best results in most metrics. On average across the random, popular, and adversarial settings, RC-DPO improves the F1 score by 2.16\% on ThinkLite-VL and 2.00\% on OpenVLThinker. Moreover, RC-DPO outperforms other baselines in most settings, especially on ThinkLite-VL, where it achieves the best F1 score under all three settings. These results further demonstrate the robustness and generalizability of RC-DPO on stronger MLRM backbones.

\begin{figure}[t]
\begin{center}
\includegraphics[width=0.85\linewidth]{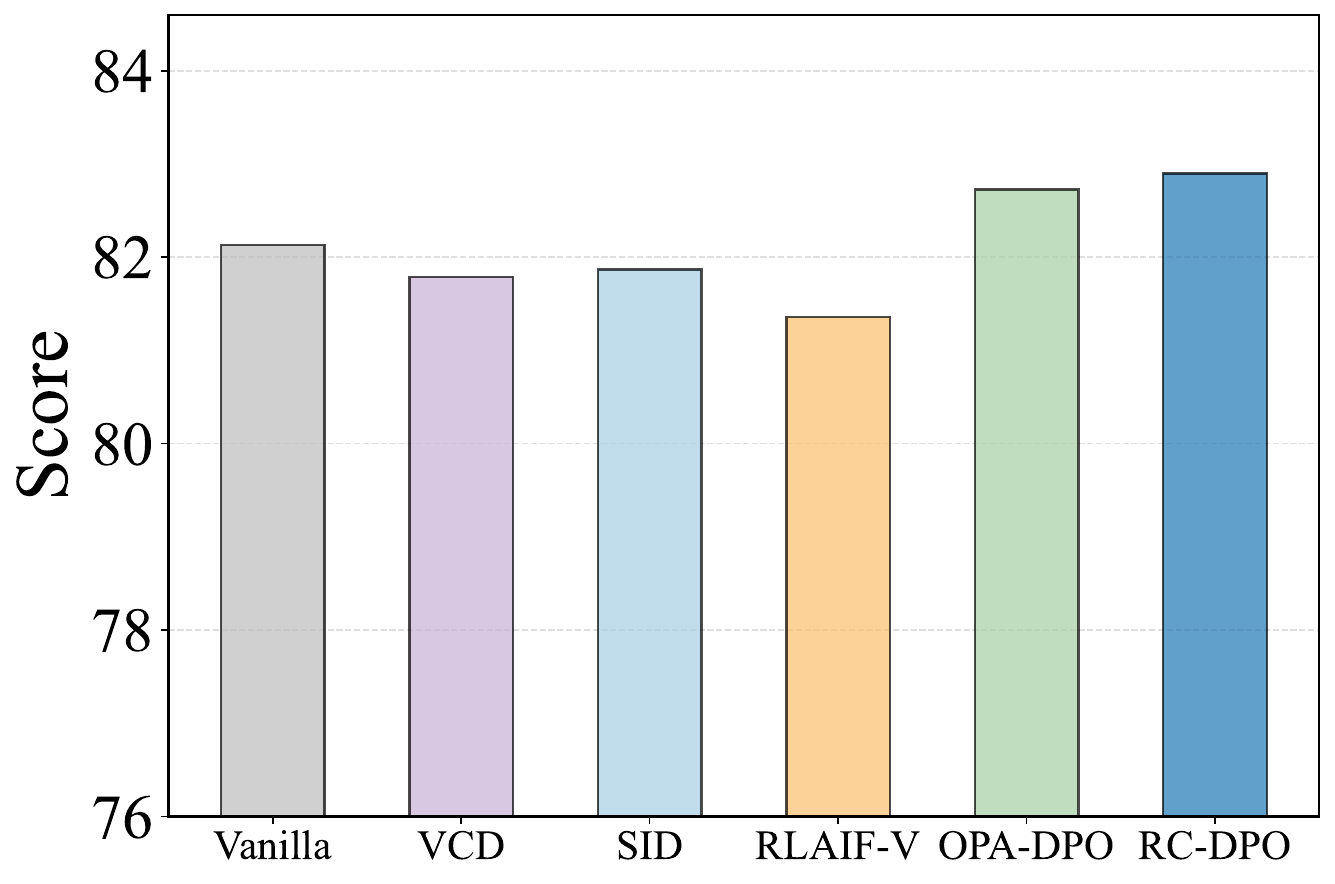}
\end{center}
\caption{Overall MMBench scores of RC-DPO and baseline methods on MM-Eureka-7B.}
\label{fig:mmbench}
\end{figure}

\subsection{Object HalBench Evaluation}
To provide a more fine-grained analysis, we decompose each generated response into CoT and final-answer segments, and compute CHAIR$_S$ for the full response, CoT, and answer separately. As shown in Figure~\ref{fig:chair}, the vanilla models exhibit substantially higher hallucination scores in the CoT than in the final answer, which largely contributes to the high hallucination rate of the full response. OPA-DPO, based on response-level preference optimization, alleviates hallucinations to some extent on several models. However, due to the answer-shortcut phenomenon discussed in Section~\ref{sec:motivation}, CoT hallucination remains significantly higher than answer hallucination, indicating that reasoning chains are still insufficiently aligned. In contrast, RC-DPO strengthens CoT-level alignment by explicitly optimizing the reasoning-conditioned preference signal. As a result, it reduces hallucinations not only in the final answer but also in the CoT segment, leading to a more consistent reduction in full-response hallucination. These results further demonstrate that explicit reasoning-chain alignment is crucial for mitigating hallucinations in MLRMs.

\begin{figure}[t]
\begin{center}
\includegraphics[width=0.85\linewidth]{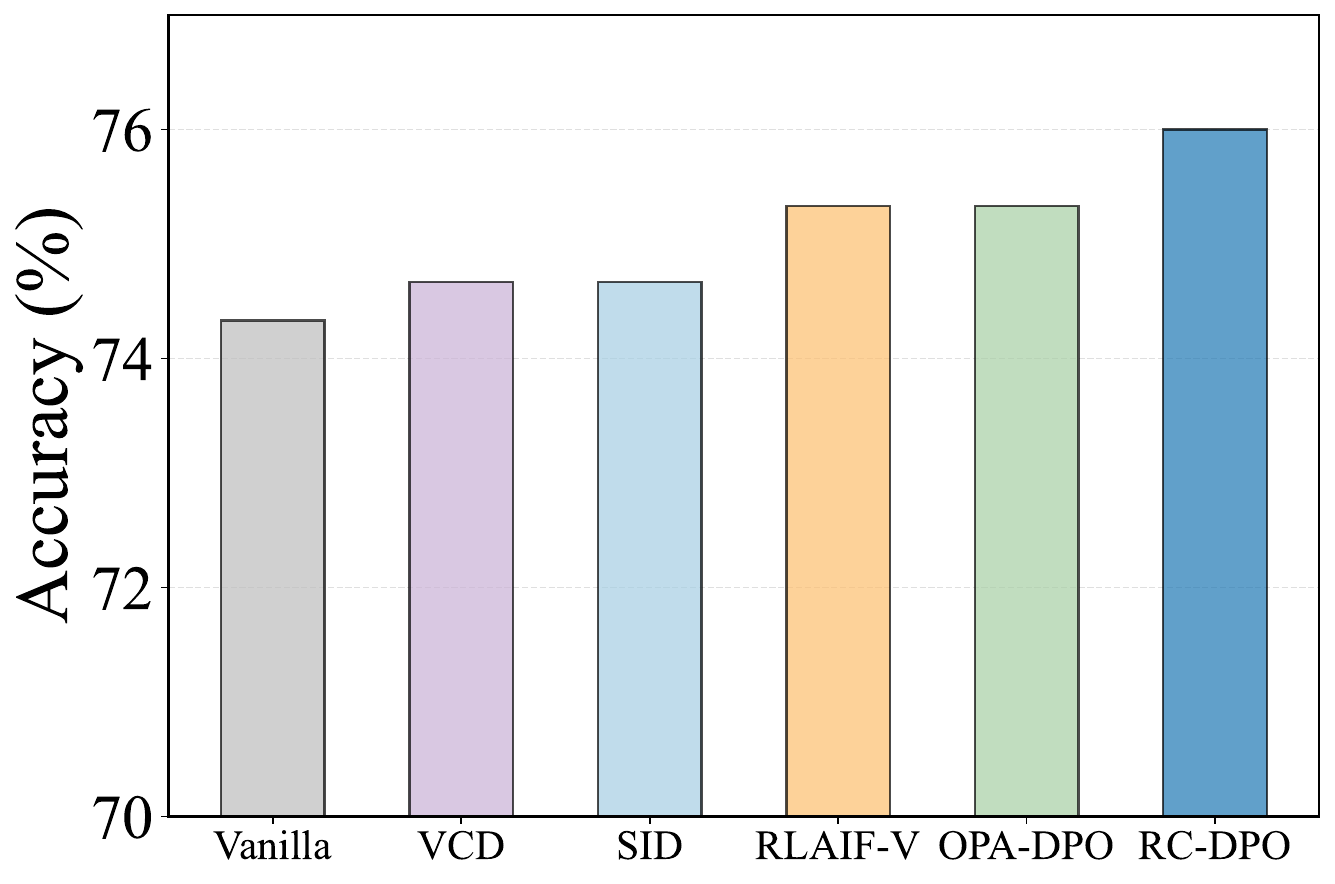}
\end{center}
\caption{MMVP accuracy of RC-DPO and baseline methods on MM-Eureka-7B.}
\label{fig:mmvp}
\end{figure}

\begin{table*}[htbp]
  \centering
  \caption{Performance comparison of RC-DPO and baseline methods on the MME benchmark with MM-Eureka.}
  \resizebox{\linewidth}{!}{
    \begin{tabular}{lcccccccccc}
    \toprule
    \multirow{2}[0]{*}{Method} & \multicolumn{5}{c}{\textbf{Perception}}              & \multicolumn{5}{c}{\textbf{Cognition}} \\
    \cmidrule(lr){2-6}
    \cmidrule(lr){7-11}
             & Total    & Position & Color    & Scene    & OCR      & Total    & Commonsense & Calculation & Translation & Code \\
             \midrule
    \textit{Vanilla}  & 1595.77  & 141.67   & 180.00   & 145.00   & 170.00   & 706.79   & 144.29   & 192.50   & 185.00   & \textbf{185.00} \\
    VCD      & 1596.14  & 131.67   & 165.00   & 148.00   & 162.50   & 715.36   & 152.86   & 200.00   & 177.50   & 185.00 \\
    SID      & 1596.38  & 135.00   & 175.00   & 146.50   & 175.00   & 705.00   & 150.00   & 192.50   & 177.50   & 185.00 \\
    RLAIF-V  & 1539.52  & 151.67   & 185.00   & 139.75   & 170.00   & 708.93   & \textbf{161.43} & 200.00   & 170.00   & 177.50 \\
    OPA-DPO  & 1557.08  & 141.67   & 180.00   & 147.50   & 147.50   & 702.86   & 147.86   & 185.00   & \textbf{192.50} & 177.50 \\
    \gc RC-DPO   & \gc\textbf{1613.03} & \gc\textbf{153.33} & \gc\textbf{185.00} & \gc\textbf{150.50} & \gc\textbf{192.50} & \gc\textbf{718.93} & \gc156.43   & \gc\textbf{200.00} & \gc185.00   & \gc177.50 \\
    \bottomrule
    \end{tabular}%
    }
  \label{tab:mme}%
\end{table*}%

\begin{table*}[htbp]
  \centering
  \caption{Performance comparison of RC-DPO and baseline methods on VMCBench with MM-Eureka}
   \resizebox{\linewidth}{!}{
    \begin{tabular}{lccccccccccc}
    \toprule
    Method   & \textbf{Overall} & \textbf{MMMU} & \textbf{MathVista} & \textbf{AI2D} & \textbf{VQAv2} & \textbf{SEEDBench} & \textbf{General} & \textbf{TextVQA} & \textbf{MMStar} & \textbf{DocVQA} & \textbf{A-OKVQA} \\
    \midrule
    \textit{Vanilla}  & 80.88    & 57.93    & 73.27    & 79.04    & 90.05    & 76.05    & 83.52    & 96.63    & 63.66    & 91.76    & 88.47 \\
    VCD      & 80.95    & 57.93    & \textbf{74.63} & 79.73    & 89.81    & 78.52    & 84.25    & 96.4     & 65.08    & 90.87    & 89.18 \\
    SID      & 80.72    & 58.89    & 74.22    & 78.59    & 89.58    & 79.26    & 83.99    & 95.51    & 61.76    & 91.09    & 88.71 \\
    RLAIF-V  & 80.85    & 57.45    & 69.31    & \textbf{80.87} & 90.51    & 79.26    & 84.21    & 96.4     & \textbf{65.8} & 95.32    & 88.71 \\
    OPA-DPO  & 81.39    & 58.89    & 73.76    & 77.9     & 90.51    & \textbf{80.74} & 84.09    & 96.85    & 64.61    & 98.44    & 89.65 \\
    \gc RC-DPO   & \gc\textbf{81.55} & \gc\textbf{59.62} & \gc74.26    & \gc80.41    & \gc\textbf{91.44} & \gc80.25    & \gc\textbf{84.73} & \gc\textbf{97.53} & \gc65.08    & \gc\textbf{98.44} & \gc\textbf{90.12} \\
    \bottomrule
    \end{tabular}%
    }
  \label{tab:vmcbench}%
\end{table*}%

\subsection{MMHal-Bench Evaluation}

MMHal-Bench evaluates hallucinations in open-ended multimodal responses. Following the standard evaluation protocol, we prompt each model to generate answers for the image-question pairs in MMHal-Bench, and then use GPT-4 to judge the response quality and hallucination severity. We report the average response score and hallucination rate, where a higher average score indicates better answer quality and a lower hallucination rate indicates fewer hallucinated contents.

As shown in Table~\ref{tab:mmhal}, RC-DPO consistently achieves the best performance across all four MLRMs. Specifically, RC-DPO obtains the highest average score and the lowest hallucination rate on four MLRMs. The improvement is particularly clear on MM-Eureka, where RC-DPO increases the average score from 3.33 to 4.01 and reduces the hallucination rate from 54.17\% to 40.62\% compared with the vanilla model. Similar trends are observed on the other models, demonstrating that RC-DPO effectively improves open-ended response faithfulness while maintaining answer quality.

\subsection{AMBER Evaluation}
We also report the results on the AMBER benchmark in Table~\ref{tab:amber}, which includes both generative and discriminative hallucination evaluations.
RC-DPO achieves consistently strong performance across the three evaluated MLRMs. In the generative setting, RC-DPO obtains the lowest CHAIR, HalRate, and Cog scores on all models, indicating that it effectively reduces hallucinated content while improving response faithfulness. Specifically, the improvements are particularly significant on R1-Onevision, where RC-DPO reduces HalRate from 35.6 to 20.8 compared with the vanilla model. In the discriminative setting, RC-DPO also achieves the best overall performance in most metrics, especially improving the F1 score from 73.87 to 88.76 on R1-Onevision. These results further demonstrate that RC-DPO generalizes well to both open-ended generation and discriminative evaluations.

\subsection{MME Evaluation}
To assess the impact of RC-DPO on general multimodal performance, we evaluate MM-Eureka on the MME benchmark, as shown in Table~\ref{tab:mme}. RC-DPO achieves the best overall performance in both perception and cognition categories. 
In the perception subset, RC-DPO consistently improves fine-grained visual understanding, obtaining the highest scores on position, color, scene, and OCR. Notably, the OCR score increases from 170.00 to 192.50 compared with the vanilla model, indicating stronger visual-text grounding after reasoning-chain alignment. 
In the cognition subset, RC-DPO also achieves the highest total score and performs competitively across multiple subtasks. These results show that RC-DPO does not sacrifice general multimodal capabilities while mitigating hallucinations. Instead, it further improves the model's perception and reasoning performance.

\subsection{MMBench Evaluation}
MMBench is a general-purpose multimodal benchmark designed to evaluate the comprehensive vision-language understanding ability of MLLMs. Following the official evaluation protocol, we prompt each model to answer multiple-choice questions and report the overall score. This benchmark allows us to examine whether hallucination mitigation affects general multimodal capability beyond hallucination-oriented evaluation.

As shown in Figure~\ref{fig:mmbench}, RC-DPO achieves the highest overall score among all compared methods. Compared with the vanilla model and existing baselines, RC-DPO brings consistent improvement, indicating that reasoning-chain alignment does not compromise general multimodal understanding. Instead, by encouraging more faithful intermediate reasoning, RC-DPO further improves the overall performance on broad vision-language tasks.

\subsection{MMVP Evaluation}

MMVP focuses on evaluating fine-grained visual perception and is designed to expose visual shortcomings of multimodal models, especially cases where models rely on language priors instead of carefully grounding their answers in the image. Following the official protocol, we evaluate each method on MMVP and report the accuracy.

As shown in Figure~\ref{fig:mmvp}, RC-DPO achieves the best accuracy among all compared methods. The improvement over the vanilla model and baseline approaches suggests that RC-DPO preserves and enhances fine-grained visual grounding ability during hallucination mitigation. These results further demonstrate that explicitly aligning reasoning chains can improve visual faithfulness without sacrificing perception-oriented capabilities.

\subsection{VMCBench Evaluation}
VMCBench offers a comprehensive evaluation of multimodal capability by covering diverse subtasks, including mathematical reasoning, diagram understanding, general VQA, text-rich image understanding, document comprehension, and knowledge-based VQA. Therefore, it provides a broader test of whether hallucination mitigation preserves general reasoning and perception abilities. 
As shown in Table~\ref{tab:vmcbench}, RC-DPO achieves the highest overall score on MM-Eureka, improving the vanilla model from 80.88 to 81.55 and outperforming all baseline methods. This gain is not driven by a single subtask; RC-DPO achieves the best results on multiple subtasks, including MMMU, VQAv2, and DocVQA, covering both perception-oriented and reasoning-oriented evaluations. These results indicate that RC-DPO improves overall multimodal performance while maintaining strong capabilities across diverse tasks.

\section{Prompt}
\label{appendix:prompt}
In this section, we present the prompts used in our experiments. Specifically, Figure~\ref{fig:few_shot_prompt} shows the few-shot prompt used to generate well-structured reasoning trajectories for MCTS-based positive sample construction, while Figure~\ref{fig:eval_prompt} illustrates the verifier prompt used for trajectory evaluation during the simulation stage of MCTS.

\begin{figure*}[htbp]
\centering
\vspace{2pt}

\begin{minipage}{0.98\textwidth}
\begin{tcolorbox}[title=Prompt, colback=white, colframe=black, fonttitle=\bfseries, boxrule=0.5pt]
You are a multimodal reasoning assistant.

You are given an image and a question about the image.
Your task is to answer the question accurately by reasoning step by step
STRICTLY based on what is visible in the image.

You MUST strictly follow the format below.
\vspace{6pt}

=====================
OUTPUT FORMAT
=====================

You MUST produce TWO sections in the following order:

<think>
REASONING CONTENT
</think>

<answer>
FINAL ANSWER CONTENT
</answer>
\vspace{6pt}

=============
REASONING FORMAT (INSIDE <think>)
=============

1. Think step by step using explicit reasoning.

2. Each reasoning step MUST:

   - Describe exactly ONE logical step
   
   - Be directly grounded in clearly observable visual evidence
   
   - Avoid unsupported assumptions or speculation
   
   - End with the special token <END>
\vspace{6pt}

===========
FINAL ANSWER FORMAT (INSIDE <answer>)
===========

After completing all reasoning steps, provide the final answer.

The final answer MUST:

- Be written as a coherent and natural paragraph

- Directly and fully answer the question

- Be fluent and precise in expression

- NOT include <END>

\vspace{6pt}

=====================
IMPORTANT RULES
=====================

- Do NOT repeat the question.

- Do NOT add extra explanations outside the specified format.

- Do NOT introduce information that is not supported by the image.

- Do NOT guess or infer details that cannot be visually verified.

- Do NOT produce any content outside <think> and <answer>.
\vspace{6pt}

=====================
EXAMPLE 1
=====================

Image: A photo shows several crafting tools on a table, including scissors, a hole punch, and cutting tools.

Question: Who is more likely to use these tools, a leather crafter or a paper crafter?

Answer:
<think>
\#\#\# Step 1: The image shows tools such as scissors, a hole punch, and cutting implements placed together. <END>
\#\#\# Step 2 ...
</think>
<answer>
...
</answer>
\vspace{6pt}

=====================
EXAMPLE 2 
=====================

Image: A woman stands on a red carpet holding a microphone, with cameras pointed toward her and brand logos visible in the background.

Question: What are the main elements in this image?

Answer:
<think>
\#\#\# Step 1: The image clearly shows a woman ... <END>
\#\#\# Step 2 ...
</think>
<answer>
...
</answer>
\vspace{6pt}

=====================
NOW ANSWER THE QUESTION
=====================
\end{tcolorbox}
\caption{Few-shot prompt used to generate well-structured reasoning trajectories for MCTS-based positive sample construction. Parts of the demonstrations are omitted due to space limitations.}
\label{fig:few_shot_prompt}
\end{minipage}
\end{figure*}

\begin{figure*}[htbp]
\centering
\vspace{2pt}

\begin{minipage}{0.98\textwidth}
\begin{tcolorbox}[title=Prompt, colback=white, colframe=black, fonttitle=\bfseries, boxrule=0.5pt]
You are a strict hallucination inspector for multimodal question answering.

\vspace{6pt}
You will be given:

- An image

- A question

- A model-generated answer

\vspace{6pt}
Important rules:

- The image is the ONLY ground truth.

- You must rely ONLY on what is clearly visible in the image.

- Do NOT assume, infer, or guess any information that is not visually evident.

- If something cannot be confirmed from the image, treat it as NOT SUPPORTED.

- Ignore fluency, style, and completeness.

- Focus ONLY on hallucinations.

\vspace{6pt}
Step 1: Decompose the model-generated answer into individual factual statements (atomic claims).
Each statement should describe a single object, attribute, action, or relationship.

\vspace{6pt}
Step 2: For each statement, judge whether it is:

- SUPPORTED by the image (clearly visible)

- NOT SUPPORTED by the image (not visible or cannot be confirmed)

- CONTRADICTED by the image (clearly inconsistent with the image)
\vspace{6pt}

Hallucination includes:

- Mentioning objects not visible in the image

- Inventing attributes, actions, or relationships not supported by the image

- Making claims that contradict the image content

- Making assumptions beyond what is directly observable (e.g., intent, identity, causality)

- Adding unnecessary or irrelevant statements that are not required to answer the question
\vspace{6pt}

Step 3: Briefly summarize:

- The number of hallucinated statements

- Their types (NOT SUPPORTED or CONTRADICTED)
\vspace{6pt}

Step 4: Assign a final score based STRICTLY on hallucination severity:

- **2**: No hallucination. All statements are directly supported by the image and relevant to the question.

- **1**: Minor issues. One or two statements are not fully supported or slightly irrelevant.

- **0**: Major hallucination. Multiple unsupported statements or any clear contradiction.
\vspace{6pt}

Output format (STRICT):

Reason: <one or two sentences explaining hallucination judgment>

Score: <0|1|2>

Do NOT output anything else.
\vspace{6pt}

=====================
INPUT
=====================

[Question]: \{\}

[Model-Generated Answer]: \{\}
\vspace{6pt}

=====================
OUTPUT
=====================
\end{tcolorbox}
\caption{Prompt used to verify the quality of CoT-answer trajectories during MCTS sampling.}
\label{fig:eval_prompt}
\end{minipage}
\end{figure*}

\begin{figure*}[t]
\begin{center}
\includegraphics[width=\linewidth]{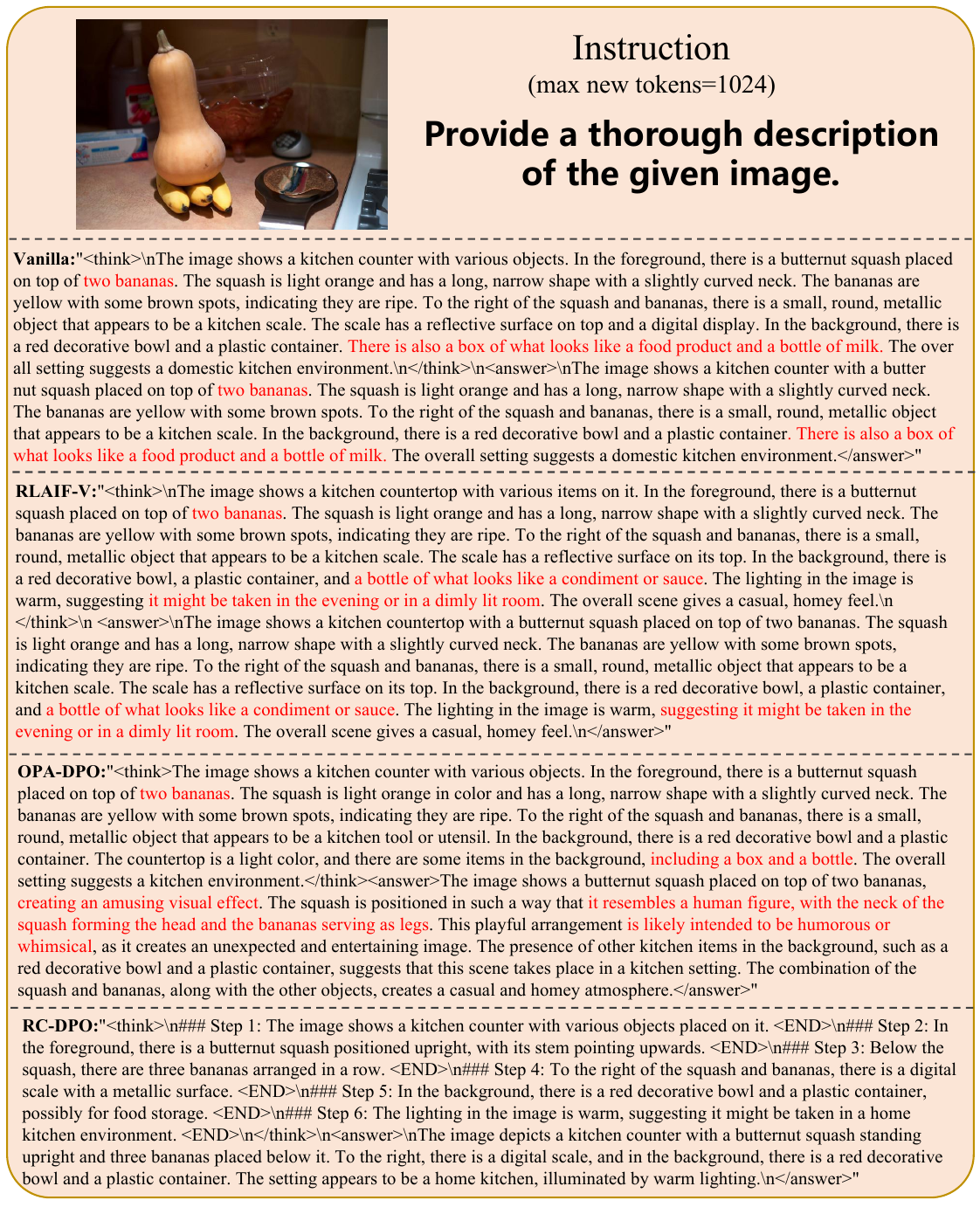}
\end{center}
\vspace{-0.5em}
\caption{Visualization results comparing our RC-DPO and other methods on MM-Eureka-7B. Hallucinations are marked in \textcolor{red}{red}.}
\label{fig:visualization}
\end{figure*}

\section{Visualization Results}

To visually demonstrate the effectiveness of our approach, we present qualitative captioning examples on Object HalBench in Figure~\ref{fig:visualization}. The results show that RC-DPO generates more visually grounded descriptions and effectively reduces hallucinations compared with baseline methods.

\end{document}